%% file: log_2022.tex
\documentclass{article}
\usepackage{log_2022}   

\usepackage{booktabs}           
\usepackage{multirow}           
\usepackage{multicol}
\usepackage{amsfonts}           
\usepackage{graphicx}           
\usepackage{duckuments}         
\usepackage{array}
\usepackage{lipsum}
\usepackage{subcaption}
\usepackage{wrapfig}

\usepackage[utf8]{inputenc} 
\usepackage[T1]{fontenc}    
\usepackage{url}            
\usepackage{nicefrac}       
\usepackage{microtype}      

\usepackage[numbers,compress,sort]{natbib}

\usepackage[mathscr]{euscript}
\usepackage[export]{adjustbox}

\usepackage{enumitem}
\newcommand{\etal}{\textit{et al.}\xspace}
\newcommand{\alg}{\textsc{PHIL}\xspace}
\newcommand{\sail}[0]{\textsc{SaIL}\xspace}

\usepackage[ruled,vlined]{algorithm2e}  
\SetKwComment{Comment}{$\triangleright$\ }{}

\usepackage{amsmath,amssymb}

\DeclareMathOperator{\EX}{\mathbb{E}} 
\newcommand{\VS}{\mathcal{V}}
\newcommand{\xhdr}[1]{{\noindent\bfseries #1}.}

\title[Learning Graph Search Heuristics]{Learning Graph Search Heuristics}

\author[M. Pándy et al.]{%
Michal Pándy\thanks{Correspondence to michalpandy@google.com.}\\
\institute{University of Cambridge}\\
\And
Weikang Qiu\\
\institute{Yale University}\\
\And
Gabriele Corso\\
\institute{MIT}\\
\And
Petar Veličković\\
\institute{DeepMind}\\
\And
Rex Ying\\
\institute{Yale University}\\
\And
Jure Leskovec \\
\institute{Stanford University} \\
\And
Pietro Liò \\
\institute{University of Cambridge} \\
}

\begin{document}

\maketitle

\begin{abstract}
Searching for a path between two nodes in a graph is one of the most well-studied and fundamental problems in computer science. In numerous domains such as robotics, AI, or biology, practitioners develop search heuristics to accelerate their pathfinding algorithms. However, it is a laborious and complex process to hand-design heuristics based on the problem and the structure of a given use case. Here we present PHIL (Path Heuristic with Imitation Learning), a novel neural architecture and a training algorithm for discovering graph search and navigation heuristics from data by leveraging recent advances in imitation learning and graph representation learning. At training time, we aggregate datasets of search trajectories and ground-truth shortest path distances, which we use to train a specialized graph neural network-based heuristic function using backpropagation through steps of the pathfinding process. Our heuristic function learns graph embeddings useful for inferring node distances, runs in constant time independent of graph sizes, and can be easily incorporated in an algorithm such as A* at test time. Experiments show that PHIL reduces the number of explored nodes compared to state-of-the-art methods on benchmark datasets by $58.5\%$ on average, can be directly applied in diverse graphs ranging from biological networks to road networks, and allows for fast planning in time-critical robotics domains.
\end{abstract}

\input{000intro}
\input{002preliminaries}
\input{001related_work}
\input{003approach}
\input{004experiments}
\input{005conclusion}
\bibliographystyle{unsrtnat}
\bibliography{reference}

\appendix
\input{006appendix}

\end{document}

%% file: 000intro.tex
\section{Introduction}
\label{sec:intro}




Search heuristics are essential in several domains, including robotics, AI, biology, and chemistry \cite{bhardwaj2017learning, chen2020retro, gebser2013domain, mac2016heuristic, sharma2017identifying, yeh2012pathway}. For example, in robotics, complex robot geometries often yield slow collision checks, and search algorithms are constrained by the robot's onboard computation resources, requiring well-performing search heuristics that visit as few nodes as possible~\cite{bhardwaj2017learning, mac2016heuristic}. In AI, domain-specific search heuristics are useful for improving the performance of inference engines operating on knowledge bases \cite{gebser2013domain, sharma2017identifying}. Search heuristics have been previously also developed to reduce search efforts in protein-protein interaction networks \cite{yeh2012pathway} and in planning chemical reactions that can synthesize target chemical products \cite{chen2020retro}. This broad set of applications underlines the importance of good search heuristics that are applicable to a wide range of problems.



\begin{figure}[!htpb]
\centering
\includegraphics[width=0.9\textwidth]{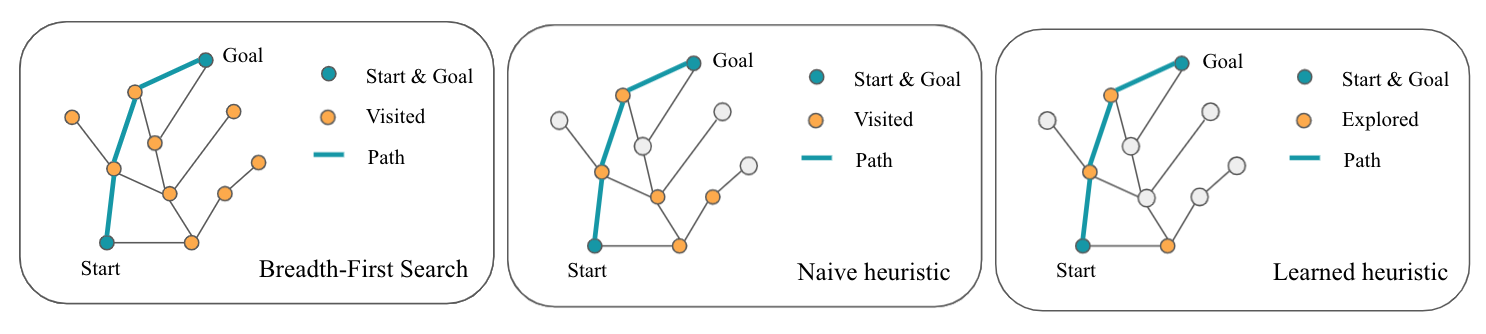}
\caption{The goal is to navigate (find a path) from the start to the goal node. While BFS visits many nodes to find a start-to-goal path (left), one can use a heuristic based on the features of the nodes (e.g., Euclidean distance) on the graph to reduce the search effort (middle). We propose \alg to learn a tailored search heuristic for a given graph, capable of reducing the number of visited nodes even further by exploiting the inductive biases of the graph (right). }
\label{fig:money-fig}
\end{figure}

The search task can be formulated as a pathfinding problem on a graph, where given a graph, the task is to navigate and find a short feasible path from a start node to a goal node, while in the process visiting as few nodes as possible (Figure~\ref{fig:money-fig}). The most straightforward approach would be to launch a search algorithm such as breadth-first search (BFS) and iteratively expand the graph from the start node until it reaches the goal node. Since BFS does not harness any prior knowledge about the graph, it usually visits many nodes before reaching the goal, which is expensive in cases such as robotics where visiting nodes is costly. To visit fewer nodes during the search, one may use domain-specific information about the graph via a \textit{heuristic} function \cite{pearl1984heuristics}, which allows one to define a distance metric on graph nodes to prune directions that seem less promising to explore. Unfortunately, coming up with good search heuristics requires significant domain expertise and manual effort.

While there has been significant progress in designing search heuristics, it remains a challenging problem. Classical approaches \cite{khalidi2020t, adabala2020multi} tend to hand-design search heuristics, which requires domain knowledge and a lot of trial and error. To alleviate this problem, there has been significant development in general-purpose search heuristics based on trading-off greedy expansions and novelty-based exploration \cite{xie2012planning, xie2014adding, lipovetzky2017best, teichteil2021boundary} or search problem simplifications \cite{bonet2001planning, zhu2003landmark, richter2010lama}. These approaches alleviate some of the common pitfalls of goal-directed heuristics, but we demonstrate that if possible, it is useful to learn domain-specific heuristics that can better exploit problem structure.

On the other hand, learning-based methods face a set of different challenges. Firstly, the data distribution is not i.i.d., as newly encountered graph nodes depend on past heuristic values, which means that supervised learning-based methods are not directly applicable. Secondly, heuristics should run fast, with ideally constant time complexity. Otherwise, the overall asymptotic time complexity of the search procedure could be increased. Finally, as the environment (search graph) sizes increase, reinforcement learning-based heuristic learning approaches tend to perform poorly \cite{bhardwaj2017learning}. State-of-the-art imitation learning-based methods can learn useful search heuristics \cite{bhardwaj2017learning}; however, these methods still rely on feature-engineering for a specific domain and do not generally guarantee a constant time complexity with respect to graph sizes.

\begin{figure}[h]
\centering
\includegraphics[width=0.9\textwidth]{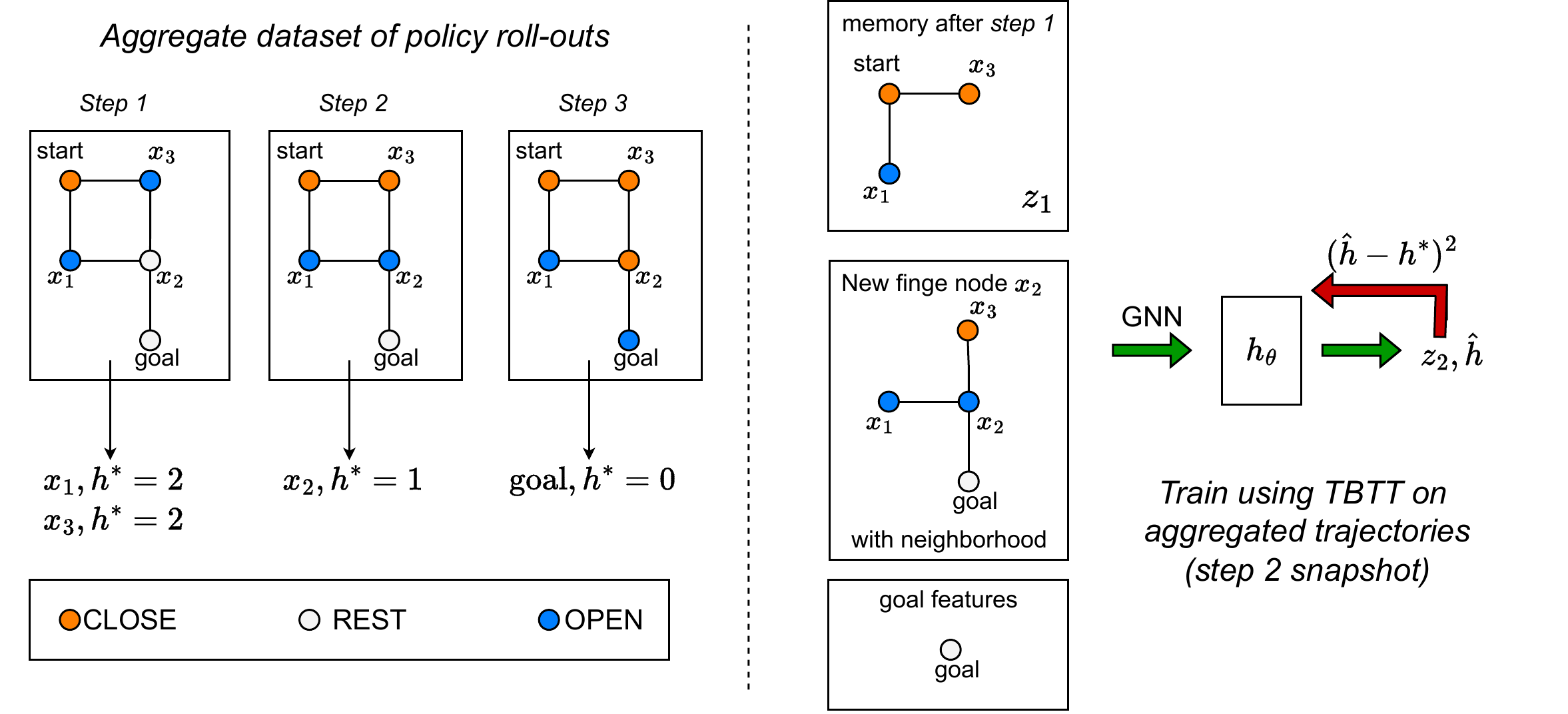}
\caption{Main components of \alg: On the left, using a greedy mixture policy induced by the current version of our parameterized heuristic $h_\theta$ and an oracle heuristic $h^*$ (i.e., a heuristic that correctly determines distances between nodes), we roll-out a search trajectory from the start node to the goal node. Each trajectory step contains a set of newly added fringe nodes with bounded random subsets of their 1-hop neighborhoods and their oracle ($h^*$) distances to the goal node. Trajectories are aggregated throughout the training procedure. On the right, we use truncated backpropagation through time on each collected trajectory to train $h_\theta$, where $\hat{h}$ is the predicted distance between $x_2$ and $x_g$, and $z_2$ is the updated state of the memory. Here, the memory captures the embedding of the graph visited so far.}
\label{fig:approach-high-level}
\end{figure}



In this paper, we propose \textit{Path Heuristic with Imitation Learning} (\alg), a framework that extends the recent imitation learning-based heuristic search paradigm with a learnable \textit{explored graph memory}. This means that \alg learns a representation that allows it to capture the structure of the so far explored graph, so that it can then better select what node to explore next (Figure \ref{fig:approach-high-level}).  We train our approach to predict the node-to-goal distances ($h^{*}$ in Figure \ref{fig:approach-high-level}) of graph nodes during search. To train our memory module, which captures the explored graph, we use truncated backpropagation through time (TBTT) \cite{sutskever2013training}, where we utilize ground-truth node-to-goal distances as a supervision signal at each search step. Our TBTT procedure is embedded within an adaptation of the AggreVaTe imitation learning algorithm \cite{ross2014reinforcement}. \alg also includes a \textit{specialized graph neural network architecture}, which allows us to apply \alg to diverse graphs from different domains.



We evaluate \alg on standard benchmark heuristic learning datasets (Section \ref{sec:experiments-grids}), diverse graph-based datasets from different domains (Section \ref{sec:experiments-diversity}), and practical UAV flight use cases (Section \ref{sec:experiments-drone}). Experiments demonstrate that \alg outperforms state-of-the-art heuristic learning methods up to $4 \times$. Further, \alg performs within $4.9 \%$ of an oracle in indoor drone planning scenarios, which is up to a $21.5 \%$ reduction compared with commonly used approaches. In practice, our contributions enable practitioners to quickly extract useful search heuristics from their graph datasets without any hand-engineering.

%% file: 002preliminaries.tex
\section{Preliminaries}
\label{sec:prelims}
\xhdr{Graph search} Suppose that we are given an unweighted connected graph $\mathcal{G} = (\VS, \mathcal{E})$, where $\VS$ is a set of nodes, and $\mathcal{E}$ a corresponding set of edges. Further suppose that each node $i \in \VS$ has corresponding features $x_i \in \mathbb{R}^{D_v}$, and each edge $(i, j) \in \mathcal{E}$ has features $e_{ij} \in \mathbb{R}^{D_e}$. Assume that we are also given a start node $v_s \in \VS$ and a goal node $v_g \in \VS$. At any stage of our search algorithm, we can partition the nodes of our graph into three sets as $\VS = \text{CLOSE} \cup \text{OPEN} \cup \text{REST}$, where CLOSE are the nodes already explored,  $\text{OPEN}$ are candidate nodes for exploration (i.e., all nodes connected to any node in CLOSE, but not yet in CLOSE), and $\text{REST}$ is the rest of the graph. Each \textit{expansion} moves a node from $\text{OPEN}$ to CLOSE, and adds the neighbors of the given node from $\text{REST}$ to $\text{OPEN}$. We call the set of newly added fringe nodes $\VS_{new}$ at each search step.  At the start of the search procedure, $\text{CLOSE} = \{v_s\}$ and we expand the nodes until $v_g$ is encountered (i.e., until $v_g \in \text{CLOSE}$).

\xhdr{Greedy best-first search} We can perform \textit{greedy best-first search} using a greedy fringe expansion policy, such that we always expand the node $v \in \text{OPEN}$ that minimizes $h(v, v_g)$. Here, $h : \VS \times \VS \longrightarrow \mathbb{R}$ is a tailored heuristic function for a given use case. In our work, we are interested in learning a function $h$ that predicts shortest path lengths, this way minimizing $|\text{CLOSE}|$ in a \textit{greedy best-first search} regime.


\xhdr{Imitation of perfect heuristics} Partially observable Markov decision processes (POMDPs) are a suitable framework to describe the problem of learning search heuristics~\cite{bhardwaj2017learning}. We can have $s = (\text{CLOSE}, \text{OPEN}, \text{REST})$ as our state, an action $a \in \mathcal{A}$ corresponds to moving a node from $\text{OPEN}$ to CLOSE, and the observations $o \in \mathcal{O}$ are the features of newly included nodes in $\text{OPEN}$. Note that one could consider an MDP framework to learn heuristics, but the time complexity of operating on the whole state is in most cases prohibitive. We also define a history $\psi \in \Psi$ as a sequence of observations $\psi = o_1, o_2, o_3, ...$. Our work leverages the observation that using a heuristic function during greedy best-first search that correctly determines the length of the shortest path between fringe nodes and the goal node will also yield minimal $|\text{CLOSE}|$. For training, we adopt a perfect heuristic $h^{*}$, similar to \cite{bhardwaj2017learning}, which has full information about $s$ during search. Such oracle can provide ground-truth distances $h^{*}(s, v, v_g)$, where $v \in \text{OPEN}$. To conclude, we define a \textit{greedy best-first search policy} $\pi_{\theta}$ that uses a parameterized heuristic $h_{\theta}$ to expand nodes from $\text{OPEN}$ with minimal heuristic values. One could also directly use a POMDP solver for the above-described problem, but this approach is usually infeasible due to the dimensionality of the search state \cite{choudhury2017adaptive}.



%% file: 001related_work.tex
\section{Related Work}

\xhdr{General purpose heuristic design} There has been significant research in designing general-purpose heuristics for speeding up satisficing planning. The first set of approaches are based on simplifying the search problem for example using landmark heuristics \cite{bonet2001planning, richter2010lama}. The next set of approaches aim to include novelty-based exploration in greedy best-first search \cite{xie2012planning, xie2014adding, lipovetzky2017best, teichteil2021boundary}. The latter set of approaches showed state-of-the-art performance (best-first width search \cite{lipovetzky2017best, teichteil2021boundary}, BFWS) in numerous settings. We show that in domains where data is available, it can be more effective to incorporate a learned heuristic into a greedy best-first search procedure.

\xhdr{Learning heuristic search} There have been numerous previous works that attempt to learn search heuristics: Arfaee \etal \cite{arfaee2011learning} propose to improve heuristics iteratively, Virseda \etal  \cite{us2013learning} learn to combine heuristics to estimate graph node distances, Wilt \etal \cite{wilt2015building} and Garrett \etal \cite{garrett2016learning} propose to learn node rankings, Thayer \etal \cite{thayer2011learning} suggest to infer heuristics during a search, and Kim \etal \cite{KimA20} train a neural network to predict graph node distances. These methods generally do not consider the non-i.i.d. nature of heuristic search. Further, Bhardwaj \etal \cite{bhardwaj2017learning} propose \sail, where heuristic learning is framed as an imitation learning problem with cost-to-go oracles. The \sail heuristic uses hand-designed features tailored for obstacle avoidance, with a linear time-complexity in the number of explored grid nodes found to be colliding with an obstacle. Feature-engineering becomes more difficult as we attempt to learn heuristics on diverse graphs such as ones seen in Section \ref{sec:experiments-diversity}, where we may need expert knowledge. Further, heuristics that do not have a constant time complexity in the size of the graph \cite{bhardwaj2017learning, ariki2019fully, terasawa20203d, yonetani2020path, archetti2021neural} generally scale poorly with graph size and hence have constrained use cases. Recent approaches to learning heuristics include Retro* \cite{chen2020retro} by Chen \etal, where a heuristic is learned in the context of AND-OR search trees for chemical retrosynthetic planning. Our work focuses on a more general graph setting.

There has been significant progress on learning heuristics for NP-hard combinatorial optimization problems \cite{NIPS2017_d9896106, NEURIPS2018_8d3bba74, NEURIPS2020_49f85a9e}. Focusing on solving NP-hard problems allows these approaches to design algorithms that are often non-exact and have a relatively large computational budget. This is not the case for methods that focus on polynomial time search, where learning-based methods are bounded by the determinism and time complexity of classical algorithms such as greedy best-first search.




\xhdr{Learning general purpose search} 
Learning general search policies is a very well-studied research area with a rich set of developments and applications. These include Monte Carlo Tree Search methods \cite{silver2010monte, guez2018learning}, implicit planning methods \cite{deac2020xlvin, karkus2017qmdp, tamar2016value}, and imagination-based planning approaches \cite{pascanu2017learning, weber2017imagination}. Learning search heuristics can be seen as a special case of \textit{general purpose search}, where the search problem is treated as a partially observable Markov decision process with restricted action evaluation (see Section \ref{sec:approach}), and with models running in $\mathcal{O}(1)$ to remain competitive time-complexity-wise on problems where best-first search performs well. \textit{General purpose search} methods do not take into account the above-mentioned constraints, which motivates the development of tailored approaches for learning heuristics \cite{ bhardwaj2017learning, chen2020retro}.


\xhdr{Imitation learning} Our approach builds on prior work in imitation learning (IL) with cost-to-go oracles. Cost-to-go oracles have been incorporated in the context of IL in methods such as SEARN \cite{daume2009search}, AggreVaTe \cite{ross2014reinforcement}, LOLS \cite{chang2015learning}, AggrevaTeD \cite{sun2017deeply}, DART \cite{laskey2017dart}, and THOR \cite{sun2018truncated}. \sail \cite{bhardwaj2017learning} presents an AggreVaTe-based algorithm for learning heuristic search. We extend \sail by incorporating a recurrent $Q$-like function, in which sense our algorithm more closely resembles AggreVaTeD by Sun \etal \cite{sun2017deeply}. While a recurrent policy can be easily incorporated in AggreVaTeD, we cannot use a policy to evaluate actions. This is due to the fact that we would either have to evaluate all actions in a state, which is computationally infeasible, or we would have to give up on taking actions that are not in the most recent version of the search fringe, which would degrade the performance (see Section \ref{sec:approach}). 


%% file: 003approach.tex
\section{Path Heuristic with Imitation Learning}
\label{sec:approach}


\xhdr{Training objective} With the aim of minimizing $|\text{CLOSE}|$ after search, our goal is to train a parameterized heuristic function $h_{\theta} : \Psi \times \VS \times \VS \longrightarrow \mathbb{R}$ to predict ground-truth node distances $h^{*}$ and use $h_\theta$ within a greedy best-first policy $\pi_\theta$ at test time.  More specifically, we assume access to a distribution over graphs $P_{\mathcal{G}}$, a start-goal node distribution $P_{v_{sg}}(\cdot \: | \: \mathcal{G})$, and a time horizon $T$. Moreover, we assume a joint state-history distribution $s, \psi \sim P_s(\cdot \: | \: \mathcal{G}, t, \pi_{\theta}, v_s, v_g)$, where $P_s$ represents the probability our search being in state $s$, at time $0 \leq t \leq T$ on graph $\mathcal{G}$ with pathfinding problem $(v_s, v_g)$, with a greedy best-first search policy $\pi_{\theta}$ using heuristic $h_{\theta}$. Hence, our goal can be summarized as minimizing the following objective:

\begin{equation}
    \label{eq:obj}
    \mathcal{L}(\theta) = \mathop{\EX}_{\substack{\mathcal{G} \sim P_{\mathcal{G}}, \\ (v_s, v_g) \sim P_{v_{sg}} \\ t \sim \mathcal{U}(0, ..., T), \\ s, \psi \sim P_s}}\big[\frac{1}{|\text{OPEN}|}\sum_{v \in \text{OPEN}} (h^{*}(s, v, v_g) - h_{\theta}(\psi, v, v_g))^{2} \big]
\end{equation}

Before we describe the algorithm that can be used to minimize $\mathcal{L}$, we rewrite $h_{\theta}$ to include a memory digest component ($z_t$), which represents an embedding of $\psi$ at time step $t$. Hence, $h_\theta$ becomes $h_{\theta} : \mathbb{R}^{d} \times \mathcal{O} \times \VS \times \VS \longrightarrow \mathbb{R}$, where $d$ is the dimensionality of our memory's embedding space. As opposed to previous methods \cite{bhardwaj2017learning}, $z_{t}$ allows us to \textit{automatically extract} relevant features for heuristic computations and concurrently \textit{reduce the computational complexity} of the heuristic function. Further, as shown in \cite{bhardwaj2017learning}, if we would use $h_{\theta}$ to evaluate all actions in a state (i.e., recalculate the heuristic values of all nodes in $\text{OPEN}$), we would need a squared reduction in the number of expanded nodes compared with BFS for \alg to bring performance benefits over BFS, which however may not be possible on all datasets. Hence, we constrain the heuristic only to evaluate new $\text{OPEN}$ nodes which we obtain after moving a node to $\text{CLOSE}$, calling the set of new fringe nodes $\VS_{new}$ after each expansion. In practice, the policy $\pi_{\theta}$  yields an algorithm equivalent to greedy best-first search, with the heuristic function replaced by $h_{\theta}$.


\subsection{Learning algorithm \& architecture}
\label{sec:ila}

\begin{algorithm}[t]
	\caption{\alg --- Sequential Heuristic Training}\label{alg:il_alg}
	Obtain hyperparameters $T$, $\beta_{0}$, $N$, $m$, $t_\tau$;\\
	Initialize $\mathcal{D} \leftarrow \emptyset$, $h_{\theta_{1}}$; \\ \label{lst:line:}
	\For{$i = 1, \ldots, N$}
	{   Sample $\mathcal{G} \sim P_{\mathcal{G}}$; \\
	    Sample $v_s, v_g \sim P_{v_{sg}}$; \\
	    Set $\beta \leftarrow \beta_0^{i}$; \\
        Set mixture policy $\pi_{mix} \leftarrow (1 - \beta) * \pi_{\theta_{i}} + \beta  * \pi^{*}$; \\
	    Collect $m$ trajectories $\tau_{ij}$ as follows; \\
		\For{$j = 1,\ldots,m$}
		{   
			Sample $t \sim \mathcal{U}(0, ..., T - t_{\tau})$;\\
			Roll-in $t$ time steps of $\pi_{\theta_{i}}$ to obtain $z_t$ and new state $s_t = (\text{CLOSE}^{0}, \text{OPEN}^{0}, \text{REST}^{0})$; \\
			Roll-out trajectory $\tau_{ij}$ as follows;\\
			\For{$k = 1,\ldots, t_{\tau}$}
		    {   
    			Update $s_{t + k - 1}$ using $\pi_{mix}$ to get new state $s_{t+k}$ and new fringe state $\text{OPEN}^{k}$; \\
    			Obtain new fringe nodes $\VS_{new} = \text{OPEN}^{k} \setminus \text{OPEN}^{k-1}$; \\
    			Update trajectory $\tau_{ij} \leftarrow \tau_{ij} \cup \{(\VS_{new}, h^{*}(s_{t+k}, \VS_{new}, v_g))\}$; \\
	        }
		    Update dataset $\mathcal{D} \leftarrow \mathcal{D} \cup \{(\tau_{ij}, z_t)\}$ or $\mathcal{D} \cup \{(\tau_{ij}, 0)\}$; \\	
	    }
	    Train $h_{\theta_{i}}$ using TBTT on each $\tau \in \mathcal{D}$ to get $h_{\theta_{i+1}}$;\\  
	}
	\Return{best performing $h_{\theta_{i}}$ on validation};
\end{algorithm}

\begin{figure}[!htpb]
    \centering
    \includegraphics[width=0.9\textwidth]{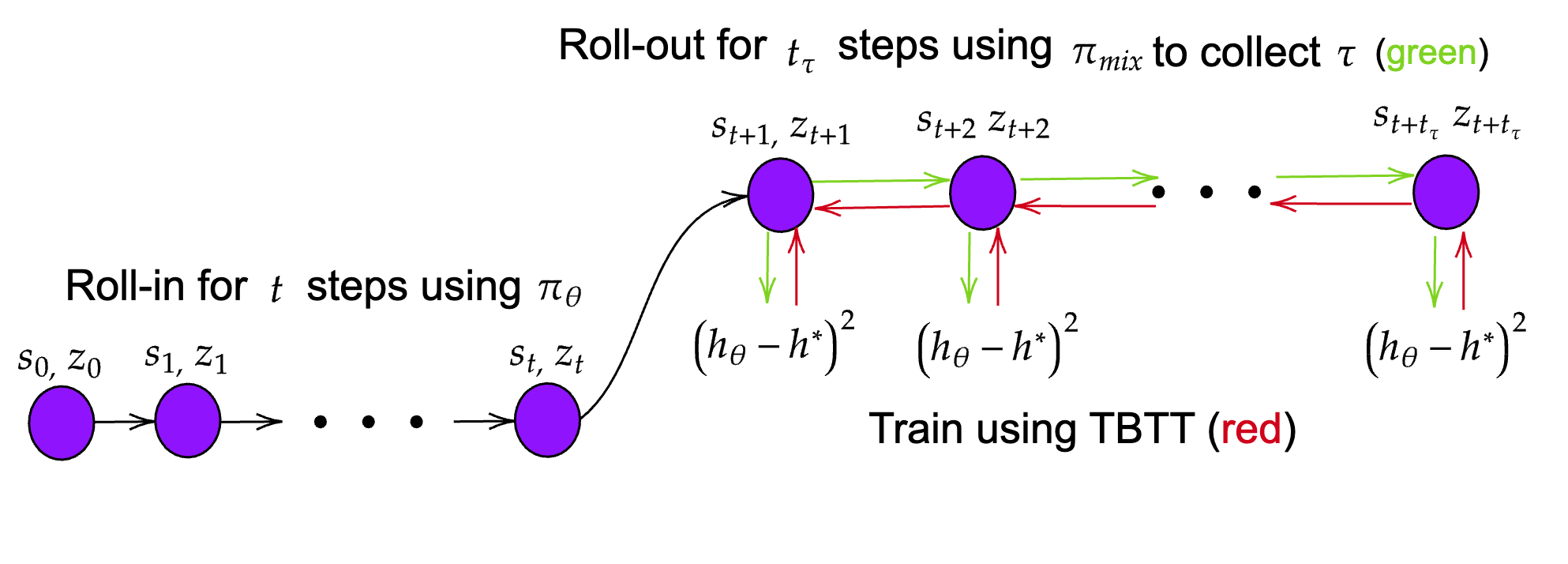}
    \caption{This figure demonstrates the core idea behind our IL algorithm. We present the roll-in phase on the left-hand side, where our policy is rolled in for $t$ steps to obtain state $s_t$ and embedding $z_t$. On the right-hand side, we show the trajectory collection and training steps, where we aggregate the trajectory for downstream training (green) and use truncated backpropagation through time on the collected dataset (red).
    }
    \label{fig:il-proc}
    
\end{figure}

\xhdr{Imitation learning algorithm} In Algorithm \ref{alg:il_alg}, we present the pseudo-code of the IL algorithm used to train our heuristic models (Figure \ref{fig:il-proc}). The high-level idea of our algorithm is that we aggregate trajectories of search traces (i.e., sequences of new fringe nodes) and use truncated backpropagation through time to optimize $h_{\theta}$ after each data-collection step. In particular, after sampling a graph $\mathcal{G}$ and a search problem  $v_s, v_g$, we use our greedy learned policy $\pi_\theta$ induced by $h_\theta$ to roll-in for $t \sim \mathcal{U}(0, \dots, T - t_{\tau})$ expansions, where $T$ is the episode time horizon, and $t_\tau$ is the roll-out length. From our roll-in, we obtain a new state $s = (\text{CLOSE}^{0}, \text{OPEN}^{0}, \text{REST}^{0})$, and an initial memory state $z_t$. After our roll-in, we roll-out for $t_{\tau}$ steps using our mixture policy $\pi_{mix}$, which is obtained by probabilistically blending $\pi_\theta$ and the greedy best-first policy induced by the oracle heuristic $\pi^{*}$. In a roll-out, we collect sequences of new fringe nodes, together with their ground-truth distances to the goal $v_g$, given by $h^{*}$.  Once the roll-out is complete, we append the obtained trajectory and the initial state for the following optimization using backpropagation through time. Further analysis on the trade-offs between using rolled-in states $z_t$ or zeroed-out states for training can be found in the supplementary material.

Note that we could also use supervised learning-based approaches to sample a fixed dataset of ($v_s$, $v_g$, $h^{*}(s, v_s, v_g)$) 3-tuples and train a model to predict node distances conditioned on their features. However, our experiments in Section \ref{sec:experiments} demonstrate that ignoring the non-i.i.d. nature of heuristic search negatively impacts model performance, with supervised learning-based methods performing up to $40 \times$ worse.


\begin{wrapfigure}[12]{R}{0.55\textwidth}
    \begin{algorithm}[H]
    	\caption{Heuristic func. ($h_\theta$) forward pass}\label{alg:htheta_alg}
    	 Obtain $x_i, x_j (j \in \mathcal{N}_i), e_{ij}, x_g$ $z_t$;\\
    	 $x_i \leftarrow f(x_i, x_g, D_{EUC}(x_i, x_g), D_{COS}(x_i, x_g)$);\\
     	 $x_j \leftarrow f(x_j, x_g, D_{EUC}(x_j, x_g), D_{COS}(x_j, x_g)$); \\
    
    	 $g_i \leftarrow \phi(x_i, \mathop{\oplus_{j \in \mathcal{N}_i}} \gamma(x_i, x_j, e_{ij}))$;\\
         $g^{\prime}_i, z_{i, t+1} \leftarrow \textsc{GRU}(g_i, z_t)$;\\
    	 $z_{t+1} \leftarrow \overline{z_{i, t+1}}$;\\ 
    	 $\hat{h}_i \leftarrow \textsc{MLP}(g^{\prime}_i, x_g);$\\
    	\Return{$\hat{h}_i ,z_{t+1}$};\\
    \end{algorithm}
\end{wrapfigure}

\textbf{Recurrent GNN architecture.} In each forward pass, $h_{\theta}$ obtains a set of new fringe nodes $\VS_{new}$, the goal node $v_g$, and the memory $z_t$ at time step $t$. We represent each node in $\VS_{new}$ using its features $x_{i} \in \mathbb{R}^{D_v}$, and likewise the goal node $v_g$ using its features $x_{g} \in \mathbb{R}^{D_v}$. Further, for each $i \in \VS_{new}$, we uniformly sample an $n \in \mathbb{N}_{\geq 0}$ bounded set of nodes present in the 1-hop neighborhood of $i$, calling this set $\mathcal{N}_i$, with $|\mathcal{N}_i| \leq n$. This sampling step produces a set of neighboring node features, where each $j \in \mathcal{N}_i$ has features $x_{j} \in \mathbb{R}^{D_v}$, and corresponding edge features $e_{ij} \in \mathbb{R}^{D_e}$.



\textbf{$h_{\theta}$ forward pass.}  Algorithm \ref{alg:htheta_alg} presents a single forward pass of $h_{\theta}$. The forward pass outputs predicted distances of the new fringe nodes to the goal $\hat{h}_i$, together with an updated memory digest $z_{t+1}$. In Algorithm \ref{alg:htheta_alg}, $f$, $\phi$, $\gamma$, $\textsc{GRU}$\cite{cho-etal-2014-learning}, $\textsc{MLP}$ are each parameterised differentiable functions, with $\phi, \gamma$ representing the \textit{update} and \textit{message} functions~\cite{gilmer2017neural} of a graph neural network, respectively.

In our forward pass, using the function $f$, we first project $x_i, x_j$ into a node embedding space, together with the goal features $x_g$, and their Euclidean ($D_{EUC}$) and cosine distances ($D_{COS}$). After that, using a $1$-layer GNN, we perform a single convolution over each $x_i$ and the corresponding neighborhood $\mathcal{N}_i$, to obtain $g_i$. The specific GNN choice is a design decision left to the practitioner, and further analysis of GNN choices can be found in Appendix D. Our graph convolution processing step allows us to easily incorporate edge features and work with variable sizes of $\mathcal{N}_i$. After the graph convolution, we apply the $\textsc{GRU}$ module over each embedding $g_i$ to obtain hidden states $z_{i, t+1}$, and new embeddings $g^{\prime}_i$. We compute the sample mean of $z_{i, t+1}$ for each node $i \in \VS_{new}$ to obtain a new hidden state $z_{t+1}$, and process $g^{\prime}_i$ with $x_g$ using an $\textsc{MLP}$ to compute the distances between the graph nodes. 

The intuitive explanation for using history embeddings is to address the partial observability of the problem. PHIL does not have access to the full graph because of time complexity concerns. The history embeddings provide a mechanism for keeping track of the belief over the full state of the graph during search. Further, the GNN allows for easily incorporating local neighborhoods and edge features. Please refer to Appendix D where we discuss the effects of our design choices.

\textbf{Permutation invariant $\VS_{new}$ embedding.} There is a trade-off between processing new fringe nodes in batch, as in Algorithm \ref{alg:htheta_alg}, and processing them sequentially. Namely, when we process the nodes in batch, we do not use the in-batch observations to predict batch node values, which means that $z_t$ is slightly outdated. On the other hand, in \alg, batch processing allows us to compute the heuristic values of all $v \in \VS_{new}$ in parallel on a GPU and preserves the memory's permutation invariance with respect to nodes in $\VS_{new}$. That is, because our observations are nodes \& edges of a graph, the respective observation ordering usually does not contain inductive biases useful for predictions, which means that we can apply a permutation invariant operator such as the mean of all new states $z_{i, t+1}$ to obtain an aggregated updated state. This approach provides additional scalability as we can process values in parallel and \alg does not have to infer permutation invariance in $ \VS_{new}$ from data.

\xhdr{Runtime complexity} Since $\forall i \in \VS_{new}: \: |\mathcal{N}_i| \leq n$, Algorithm \ref{alg:htheta_alg} together with neighborhood sampling runs in up to $nc_1 + (n+1)c_2$ operations per each node $i \in \VS_{new}$, which is $\mathcal{O}(1)$ with respect to the size of the graph. Here, $c_1$ is the maximal number of operations associated with evaluating a node, such as performing robot collision checks in dynamically constructed graphs, and $c_2$ is the maximal count of total model operations (e.g., $f$ \& $\gamma$ operations) on the node set $\{i\} \cup \mathcal{N}_i$. Note that for this analysis, we assume that $c_1$ is bounded. In general, we expect to learn a better search heuristic with increasing $n$, but in some use cases, $c_1$ may dominate overall complexity, which means the hyperparameter $n$ is helpful for practitioners to tune trade-offs between constant factors and search effort minimization.

%% file: 004experiments.tex
\section{Experiments}
\label{sec:experiments}
In our experiments, we evaluate \alg both on benchmark heuristic learning datasets \cite{bhardwaj2017learning} (Section \ref{sec:experiments-grids}) as well on a diverse set of graph datasets (Section \ref{sec:experiments-diversity}). Finally, we show that \alg can be applied to efficient planning in the context of drone flight (Section \ref{sec:experiments-drone}). Our main goal is to assess how \alg compares to baseline methods in terms of necessary expansions before the goal node is reached. Please refer to the supplementary material for information about baselines, an ablation study, and additional experiment details.



\subsection{Heuristic search in grids}
\label{sec:experiments-grids}

\begin{wrapfigure}[13]{r}{0.3\textwidth}
  \includegraphics[width=\linewidth]{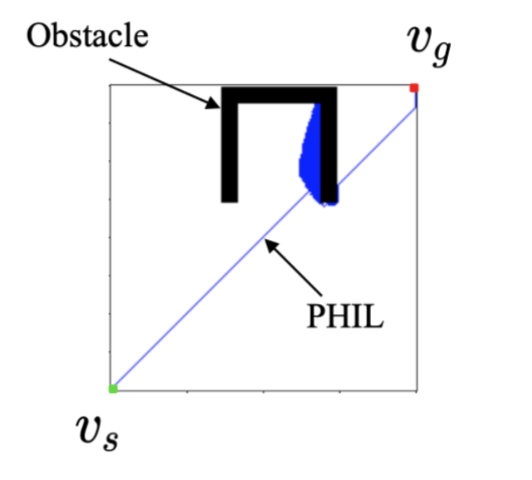}
    \caption{Example of \alg escaping local search minima.}
  \label{fig:local-min}
\end{wrapfigure}

In Section \ref{sec:experiments-grids}, we evaluate \alg on $8$, $200 \times 200$ 8-connected grid graph-based datasets by Bhardwaj \etal \cite{bhardwaj2017learning}. These datasets present challenging obstacle configurations for naive greedy planning heuristics, especially when $v_s$ is in the bottom-left of the grid, and $v_g$ in the top-right. Each dataset contains 200 training graphs, 70 validation graphs, and 100 test graphs. Example graphs from each dataset can be found in Table \ref{tab:table-1}.

We train \alg with a hyperparameter configuration of $T = 128$,  $t_{\tau} = 32$, $\beta_{0} = 0.7$, $n = 8$, and using rolled-in $z_t$ states as initial states for training. We use a 3-layer $\textsc{MLP}$ of width 128 with \textit{LeakyReLU} activations, followed by a DeeperGCN \cite{li2020deepergcn} graph convolution with \textit{softmax} aggregation. Our memory’s embedding dimensionality is $64$. The node features are 2D grid coordinates. See Appendix \ref{sec:baselines-and-datasets} for an overview of our baselines and datasets. Note that although using positional representations \cite{you2019position} might be useful for this search problem, it is not scalable as the graph size increases (we refer the reader to Appendix \ref{sec:future-work} for more details). 

\begin{table*}[!htb]
\centering
\resizebox{\columnwidth}{!}{%

\begin{tabular}{ccccccccccccccc} \toprule
 \textbf{Dataset} & \multicolumn{3}{c}{\textbf{Graph Examples}} & \textbf{\sail} & \textbf{SL} & \textbf{CEM} & \textbf{QL} & \textbf{$h_{euc}$} & \textbf{$h_{man}$} & \textbf{A*} & \textbf{MHA*} & \textbf{BFWS} & \textbf{Neural A*} & \textbf{\alg} \\ \midrule

Alternating gaps    & \frame{\includegraphics[width=0.6cm, height=0.6cm]{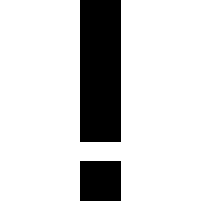}} &  \frame{\includegraphics[width=0.6cm, height=0.6cm]{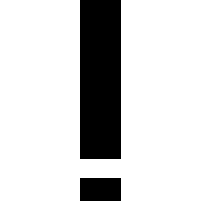}}  &  \frame{\includegraphics[width=0.6cm, height=0.6cm]{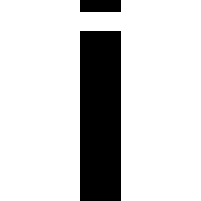}}   &0.039	&0.432& 0.042	& 1.000 & 1.000	&1.000	&1.000 & 1.000 & 0.34 &0.546 &\textbf{0.024}\\ 
Single Bugtrap       & \frame{\includegraphics[width=0.6cm, height=0.6cm]{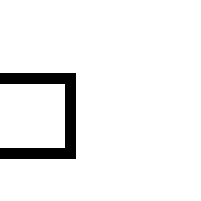}} & \frame{\includegraphics[width=0.6cm, height=0.6cm]{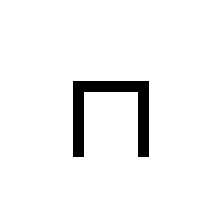}}  & \frame{\includegraphics[width=0.6cm, height=0.6cm]{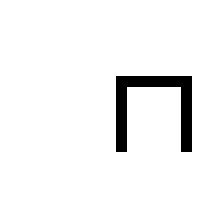}} 	& 0.158	& 0.214	& 0.057	& 1.000	&0.184	&0.192	&1.000	&0.286 & 0.099 &0.394 & 
\textbf{0.077} \\ 
Shifting gaps       & \frame{\includegraphics[width=0.6cm, height=0.6cm]{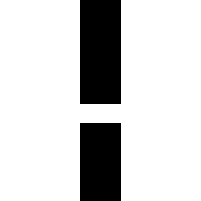}}  & \frame{\includegraphics[width=0.6cm, height=0.6cm]{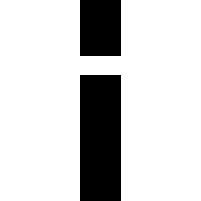}}  & \frame{\includegraphics[width=0.6cm, height=0.6cm]{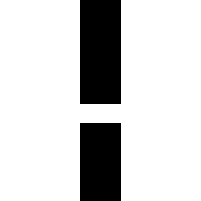}}  &0.104	&0.464	&1.000	&1.000	&0.506	&0.589	&1.000	&0.804 & 0.206 &0.563 &\textbf{0.027} \\
Forest              & \frame{\includegraphics[width=0.6cm, height=0.6cm]{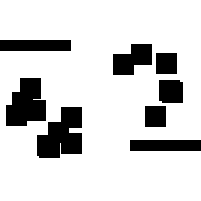}}  & \frame{\includegraphics[width=0.6cm, height=0.6cm]{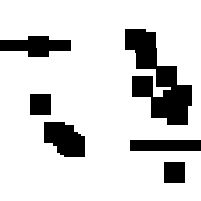}}  &  \frame{\includegraphics[width=0.6cm, height=0.6cm]{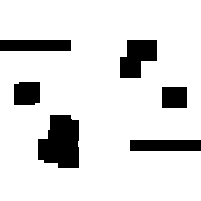}}   & 0.036 &0.043	&0.048	&0.121	&0.041	&0.043	&1.000	&0.075 & 0.039  &0.399 & \textbf{0.027}\\
Bugtrap+Forest      & \frame{\includegraphics[width=0.6cm, height=0.6cm]{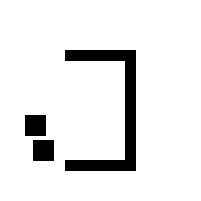}}  & \frame{\includegraphics[width=0.6cm, height=0.6cm]{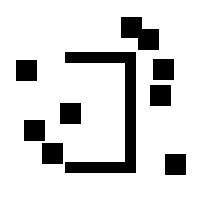}}  &  \frame{\includegraphics[width=0.6cm, height=0.6cm]{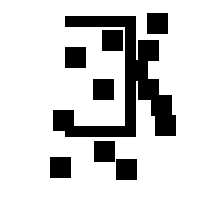}}  	&0.147	&0.384	&0.182	&1.000	&0.410	&0.337	&1.000	&3.177 & 0.149 &0.651 &\textbf{0.135} \\
Gaps+Forest         &\frame{ \includegraphics[width=0.6cm, height=0.6cm]{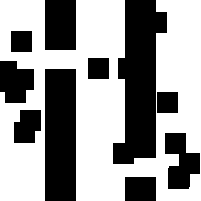}}  & \frame{\includegraphics[width=0.6cm, height=0.6cm]{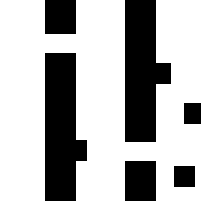}}  & \frame{\includegraphics[width=0.6cm, height=0.6cm]{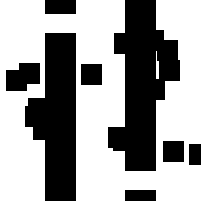}}    	&0.221	&1.000	&1.000	&1.000	&1.000	&1.000	&1.000	&1.000 & 0.401 & 0.580 &\textbf{0.039} \\
Mazes               & \frame{\includegraphics[width=0.6cm, height=0.6cm]{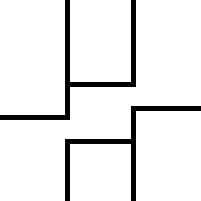}}  & \frame{\includegraphics[width=0.6cm, height=0.6cm]{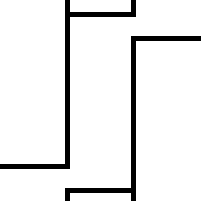}}  & \frame{\includegraphics[width=0.6cm, height=0.6cm]{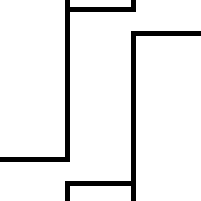}} 	&0.103	&0.238	&0.479	&0.399	&0.185	&0.171	&1.000	&0.279 & 0.095 &1.000 &\textbf{0.069}\\
Multiple Bugtraps   & \frame{\includegraphics[width=0.6cm, height=0.6cm]{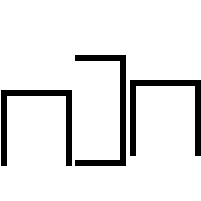}}  & \frame{\includegraphics[width=0.6cm, height=0.6cm]{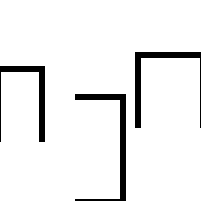}}  & \frame{\includegraphics[width=0.6cm, height=0.6cm]{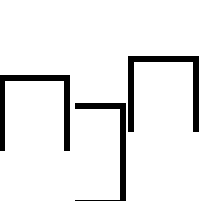}} 	&0.479	&0.480	&1.000	&0.835	&0.648	&0.617	&1.000	&0.876 &0.169 &0.331 & \textbf{0.136}  \\ \bottomrule

\end{tabular}
}
\caption{The number of expanded graph nodes of \alg with respect to \sail. We can observe that out of all baselines, \sail performs best. \alg outperforms \sail by $58.5 \%$ on average over all datasets, with a maximal search effort reduction of $82.3 \%$ in the \textit{Gaps+Forest} dataset.}
\label{tab:table-1}

\end{table*}

\begin{figure*}[h!]
    \centering
    \begin{adjustbox}{minipage=\linewidth,scale=0.9}

    \begin{subfigure}[t]{0.45\textwidth}
         \raisebox{-\height}{\includegraphics[width=0.49\textwidth,height=0.49\textwidth]{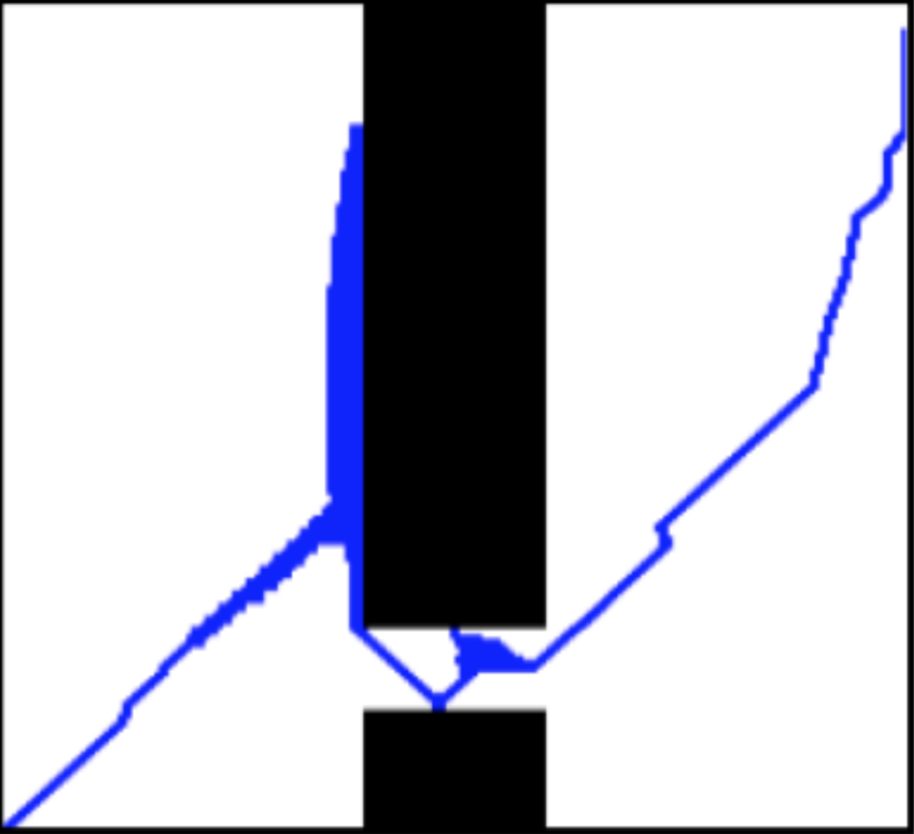}}
         \raisebox{-\height}{\includegraphics[width=0.49\textwidth,height=0.49\textwidth]{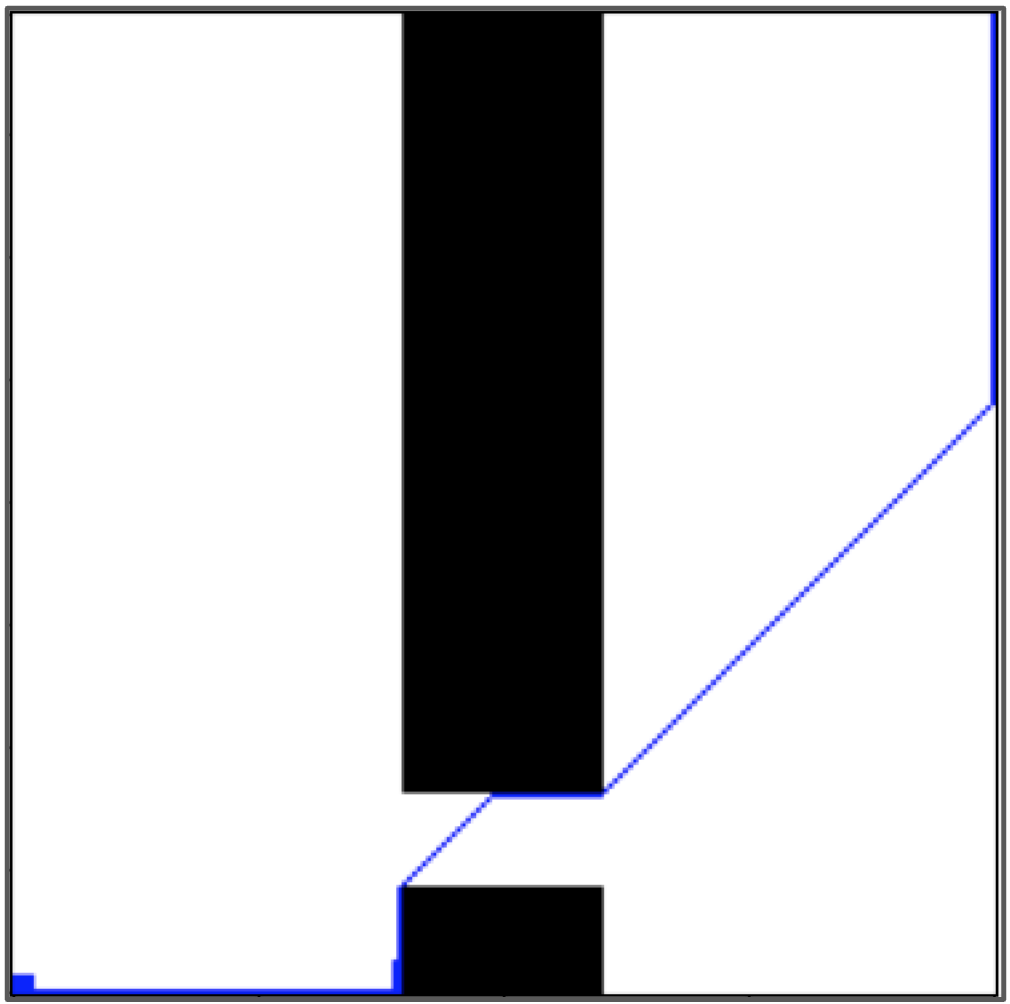}}%
         \label{fig:one-a}
    \caption{Shifting gaps (\sail left, \alg right)}
    \end{subfigure}
    \hfill
    \begin{subfigure}[t]{0.45\textwidth}
         \raisebox{-\height}{\includegraphics[width=0.49\textwidth,height=0.49\textwidth]{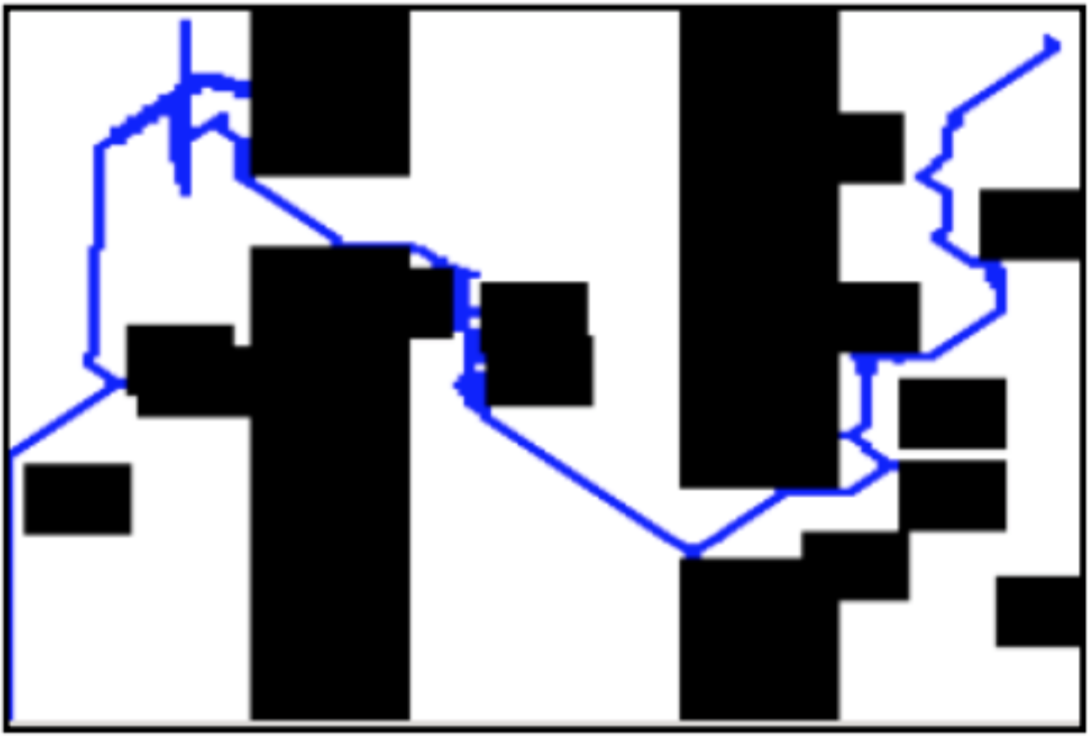}}
         \raisebox{-\height}{\includegraphics[width=0.49\textwidth,height=0.49\textwidth]{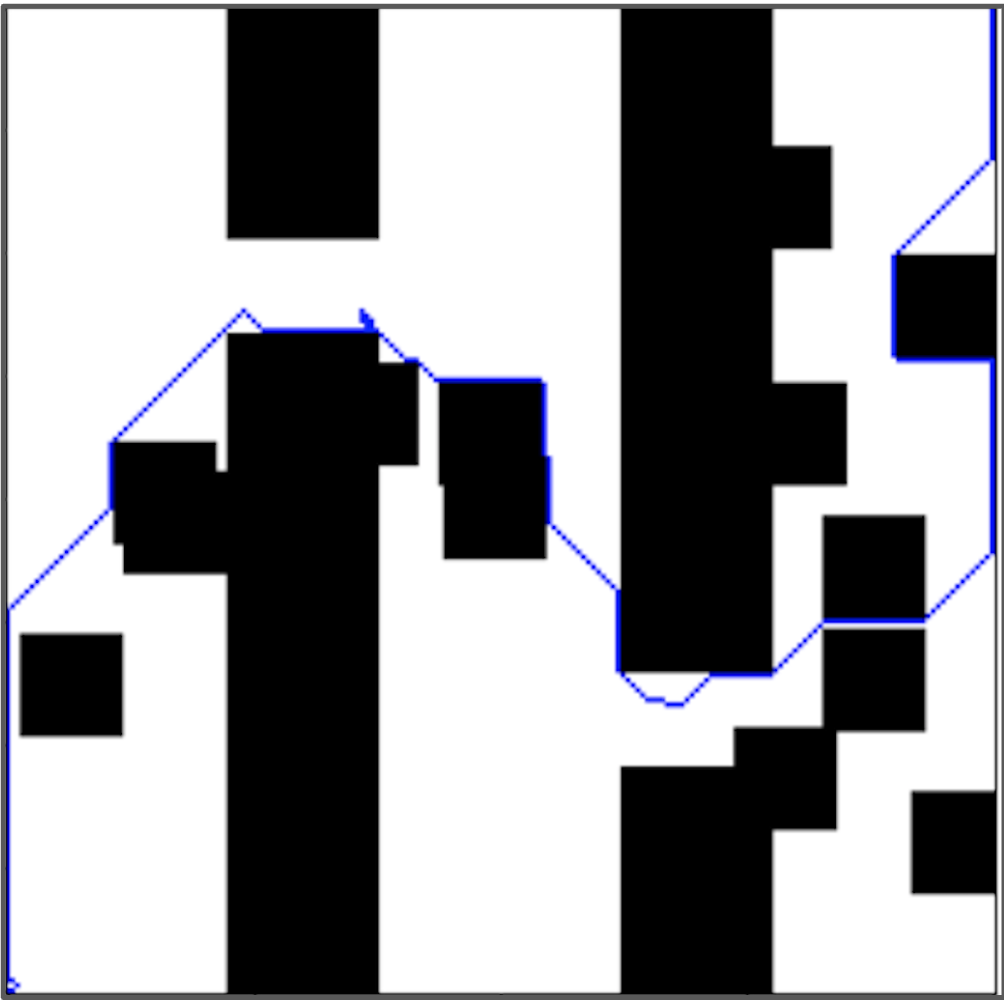}}%
    \caption{Gaps+Forest (\sail left, \alg right)}
    \end{subfigure}
    \begin{subfigure}[t]{0.45\textwidth}
        \raisebox{-\height}{\includegraphics[width=0.49\textwidth,height=0.49\textwidth]{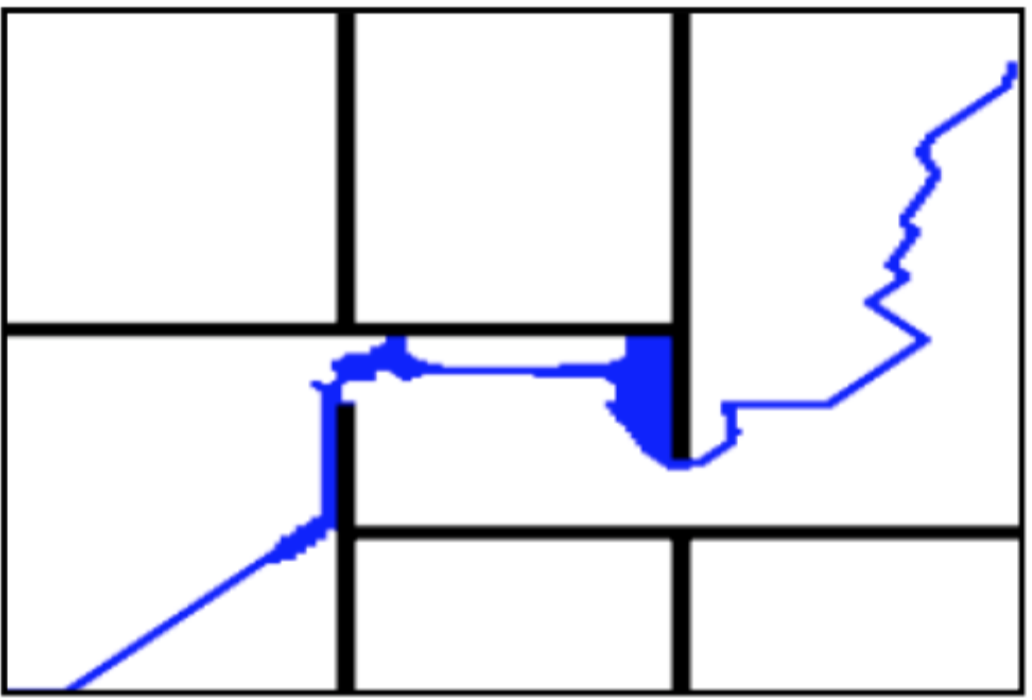}}
        \raisebox{-\height}{\includegraphics[width=0.49\textwidth,height=0.49\textwidth]{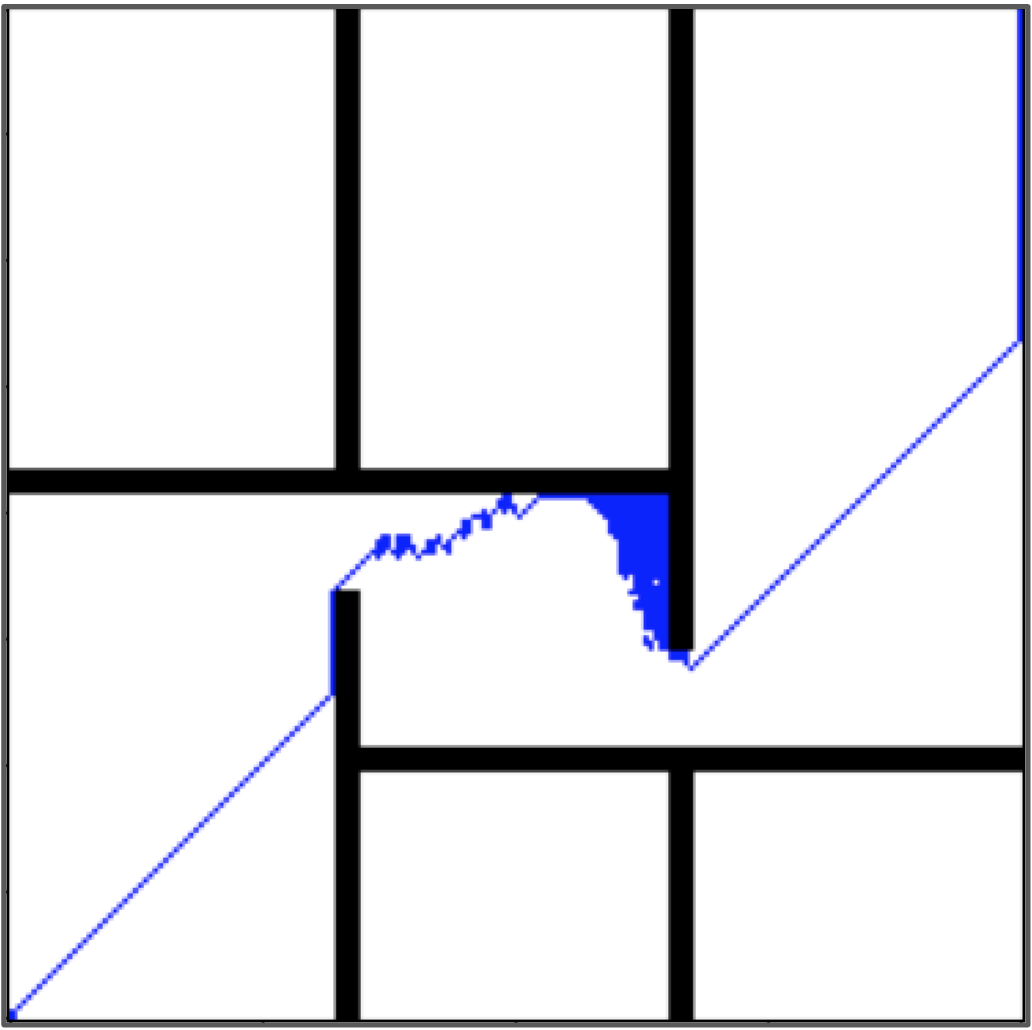}}%
    \caption{Mazes (\sail left, \alg right)}
    \end{subfigure}
    \hfill
    \begin{subfigure}[t]{0.45\textwidth}
          \raisebox{-\height}{\includegraphics[width=0.49\textwidth,height=0.49\textwidth]{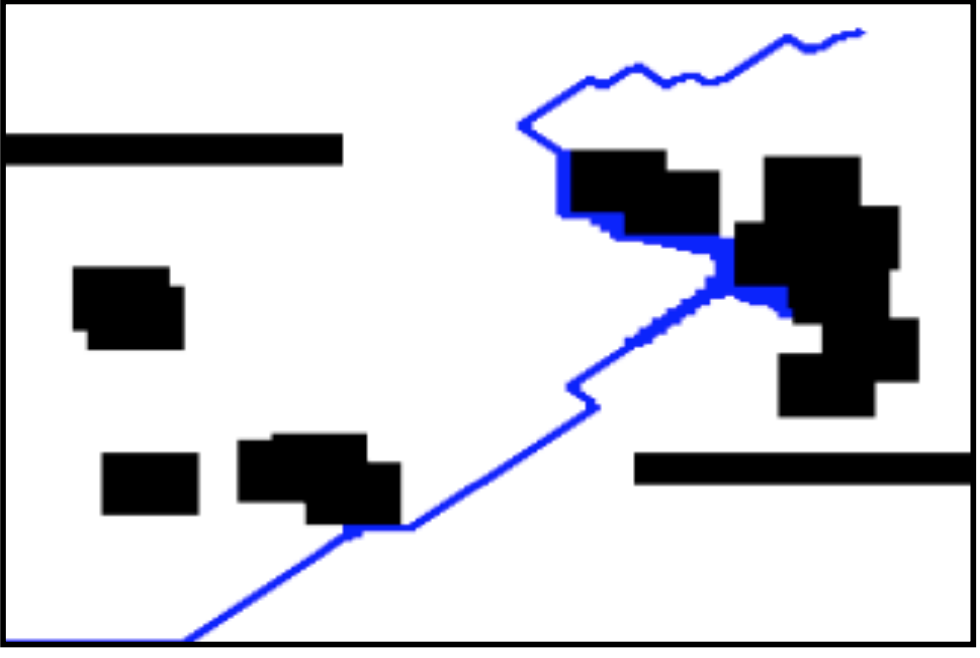}}
          \raisebox{-\height}{\includegraphics[width=0.49\textwidth,height=0.49\textwidth]{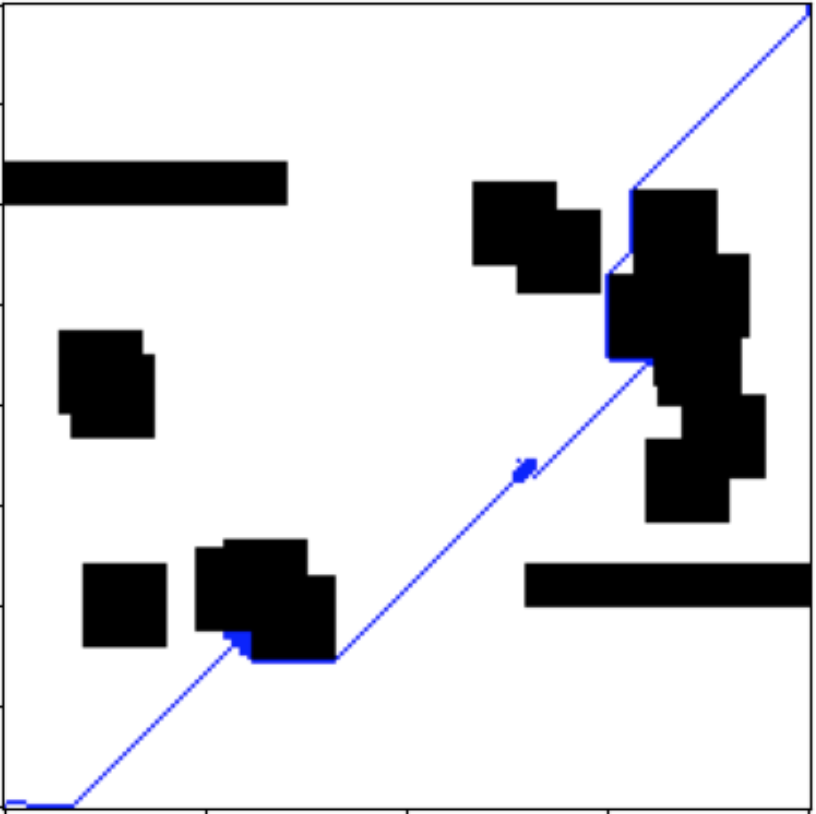}}%
    \caption{Forest (\sail left, \alg right)}
    \end{subfigure}
    \end{adjustbox}
    \caption{In each image pair of this figure, we provide a qualitative comparison with the \sail method. In particular, we show comparisons on the \textit{Shifting gaps}, \textit{Gaps+Forest}, \textit{Mazes}, and \textit{Forest} datasets. We can observe that \alg (right) learns the appropriate heuristics for the given dataset and makes fewer redundant expansions than \sail (left).
    } 
    \label{fig:sail-comparison}
\end{figure*}

\textbf{Discussion.} As we can see in Table \ref{tab:table-1}, \alg outperforms the best baseline (\textit{\sail}) on all datasets, with an average reduction of explored nodes before $v_g$ is found of $58.5 \%$.
Qualitatively, observing Figure \ref{fig:sail-comparison}, we can attribute these results to \alg's ability to \textbf{reduce the redundancy in explored nodes} during a search. Further, \alg is also \textbf{capable of escaping local minima}, which is illustrated in Figure \ref{fig:local-min}. However, note that we occasionally observe failure cases in practice, where \alg gets stuck in a bug trap-like structure. We discuss possible remedies and opportunities for future work in the supplementary material.

\xhdr{Runtime \& convergence speed} \alg converges in up to $N = 36$ iterations, with $m = 1, t_\tau = 32$ (i.e., after observing less than $N * t_\tau * max(|\VS_{new}|) \approx 9,216$ shortest path distances, where we take $max(|\VS_{new}|) = 8$ as the maximal size of $\VS_{new}$). According to figures reported in \cite{bhardwaj2017learning}, this is approximately $5\times$ less data than it takes for \sail to converge.


\subsection{Search in real-life graphs of different structures}
\label{sec:experiments-diversity}

In this experiment, our goal is to demonstrate the general applicability of \alg to various graphs. We train \alg on $4$ different groups of graph datasets: citation networks, biological networks, abstract syntax trees (ASTs), and road networks. We use the same graph for citation networks and road networks for training and evaluation, and we use $100$ random $v_s, v_g$ pairs for testing. In the case of biological networks and ASTs, we usually have train/validation/test splits of $80/10/10$, and in the case of the OGB \cite{hu2020open} datasets, we use the provided splits.

\begin{table*}[!htb]
\centering
\resizebox{\columnwidth}{!}{

\begin{tabular}{c|c|ccc|ccccccc} 
\toprule
& \textbf{Dataset} &\textbf{$|\mathcal{D}|$} &\textbf{$\bar{|\VS|}$} &\textbf{$\bar{|\mathcal{E}|}$}  & \textbf{SL} &\textbf{A*} &$h_{euc}$ &\textbf{BFS} & \textbf{\sail} & \textbf{BFWS} &\textbf{\alg} \\ \midrule
\multirow{5}{*}{Citation Networks} &Cora (Sen \etal \cite{sen2008collective}) &1 &2,708 &5,429 &2.201	&2.067	&1.000	&4.001 & 0.669 &1.378    &\textbf{0.475} \\ 
&PubMed (Sen \etal \cite{sen2008collective}))   &1 &19,717 &44,338            &2.157 &2.983	&1.000	& 3.853 & 1.196 &  1.000& \textbf{0.745}  \\ 
&CiteSeer (Sen \etal \cite{sen2008collective})) &1 &3,327 &4,732            &1.636 &1.487	&1.000	&2.190 & 1.062     &0.951  &\textbf{0.599}\\
&Coauthor (cs) (Schur \etal \cite{shchur2018pitfalls}) &1 &18,333 &81,894       &1.571	&1.069	&1.000	&2.820 & 1.941 &    1.026 &\textbf{0.835}\\
&Coauthor (physics) (Schur \etal \cite{shchur2018pitfalls}) &1 &34,493 &247,962  &4.076	&1.081	&1.000	&4.523 & $-$ & 1.012 &\textbf{0.964} \\ \midrule
\multirow{4}{*}{Biological Networks} &OGBG-Molhiv  (Hu \etal \cite{hu2020open}) &41,127 &25.5	&27.5  &1.086 &1.065 &\textbf{1.000} & 1.267 & 1.104 & 1.146 & 1.016  \\
&PPI (Zitnik \etal \cite{zitnik2017predicting})         &24 &2,372.67 &34,113.16 &0.772 &0.831 &1.000 &5.618 & 1.746 & 3.941 &\textbf{0.658}\\
&Proteins (Full) (Morris \etal \cite{morris2020tudataset})  &1,113 &39.06 &72.82 &0.995 &0.997 &1.000 &2.645 & 0.891 & 0.966 &\textbf{0.831}\\
&Enzymes (Morris \etal \cite{morris2020tudataset})            &600 &32.63 &62.14 &1.073	&1.007 &1.000 &1.358 & 1.036 & 0.992 &\textbf{0.757}\\ \midrule
\multirow{1}{*}{ASTs} &OGBG-Code2 (Hu \etal \cite{hu2020open}) &452,741 &125.2 &124.2  	&1.196  &1.013	&1.000	&1.267 & 1.029 & \textbf{0.817} & 1.219 \\ \midrule
\multirow{2}{*}{Road Networks} &OSMnx - Modena (Boeing \cite{boeing2017osmnx}) &1 &29,324 &38,309  &2.904 	&3.085	&1.000	&3.493 & 1.182 &0.997&\textbf{0.489}\\ 
&OSMnx - New York (Boeing \cite{boeing2017osmnx}) &1 &54,128 &89,618 & 39.424  &36.529	&1.000	&63.352 & 1.583&1.013 &\textbf{0.962}\\ 
\bottomrule
\end{tabular}
}
\caption{Comparison of \alg with baseline approaches on $4$ groups of datasets: citation networks, biological networks, abstract syntax trees, and road networks. "$-$" denotes being out of a 4-day's training time limit. We can observe that, on average across all datasets, \alg outperforms the best baseline per dataset by $13.4 \%$. Discounting the OGBG datasets, this number becomes $19.5 \%$.}
\label{tab:table-2}
\end{table*}


Similarly as in Section \ref{sec:experiments-grids}, our $\textsc{MLP}$ has four layers of width $128$ with \textit{LeakyReLU} activations and we use a DeeperGCN~\cite{li2020deepergcn} graph convolution with \textit{softmax} aggregation. The utilized node and edge features are the provided features in each dataset, except for a few minor modifications which are discussed in Appendix A \& Appendix C. We train an $\textsc{MLP}$ of depth $5$ and width $256$ using supervised learning (SL) for our learning-based baseline method. 


\xhdr{Discussion} The results presented in Table \ref{tab:table-2} suggest that \alg can learn superior search heuristics compared with baseline methods, outperforming top baselines per dataset in terms of visited nodes during a search by $13.4 \%$ on average. Two datasets where \alg fell short compared to other baselines are the \textit{OGBG-Molhiv} and \textit{OGBG-Code2} datasets. The \textit{OGBG-Code2} dataset adopts a \textit{project split} \cite{allamanis2019adverse} and OGBG-Mohliv adopts a \textit{scaffold split} \cite{wu2018moleculenet}, both of which ensure that graphs of different structure are present in the training \& test sets. Although \alg improved upon uninformed search (BFS) in the OGB datasets, structural graph consistency is explicitly discouraged in the above-mentioned OGBG splits. Without the OGBG datasets, \alg improves on the top baselines per dataset by $19.5 \%$ on average, and upon the Euclidean node feature heuristic ($h_{euc}$) by $20.4 \%$. Note that we trained \alg up to $N = 60$ iterations, which means that it only encountered a small subset of the pathfinding problems in the single graph setting, which means that \alg had to generalize to learn useful heuristics. Even in Cora, the $|\mathcal{D}| = 1$ dataset with least number of nodes, \alg observed roughly $6,000$ node distances during training, which is less than $0.2 \%$ of total distances in the Cora graph.

\subsection{Planning for drone flight}
\label{sec:experiments-drone}


\begin{wrapfigure}[17]{r}{0.4\textwidth}
  \includegraphics[width=0.9\linewidth]{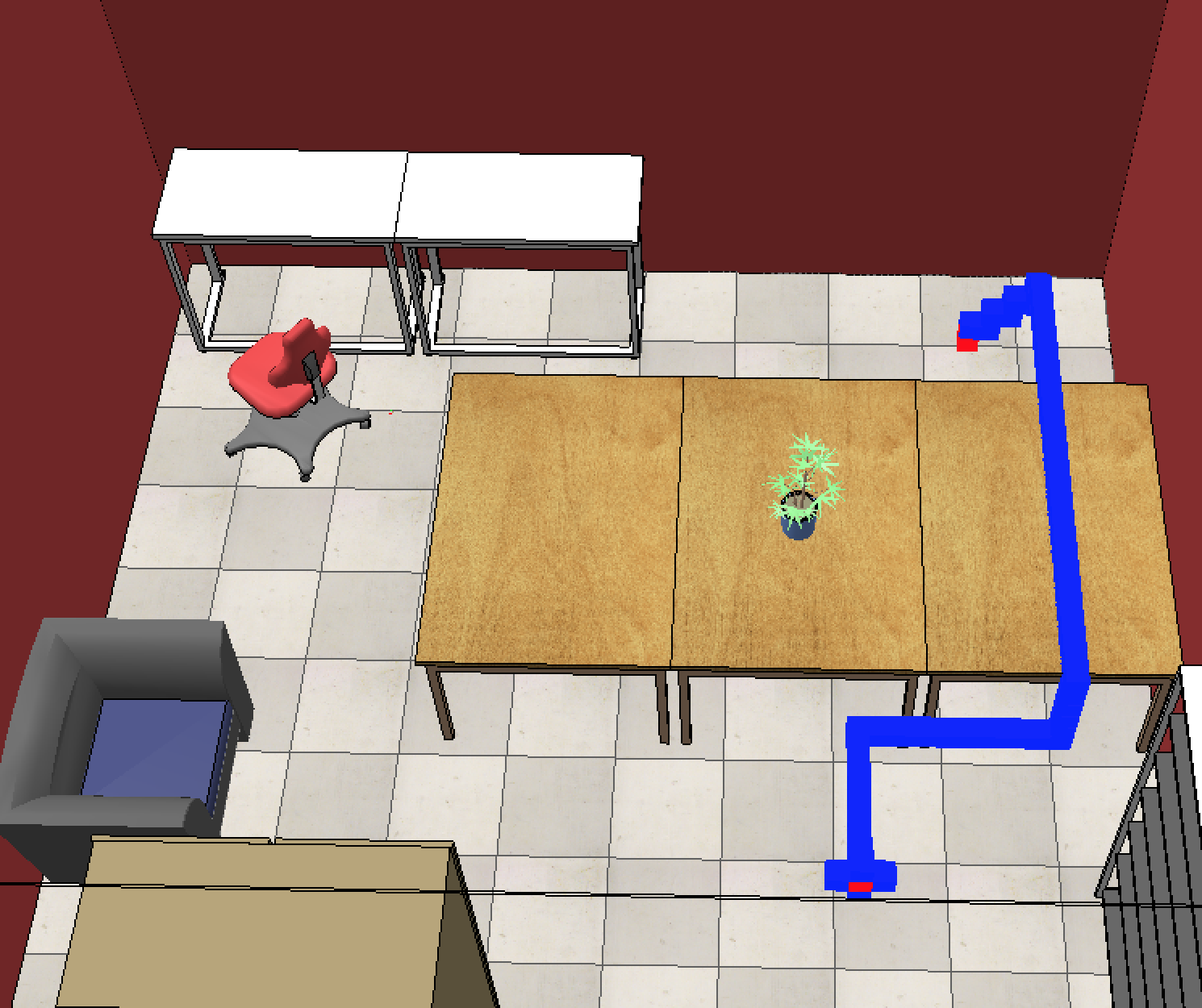}
    \caption{This figure illustrates the \textit{room adversarial} environment with an example planning problem (red) and the expanded graph by \alg (blue).}
  \label{fig:room-adv}
\end{wrapfigure}

In our final experiment, we use \alg to plan collision-free paths in a practical drone flight use case within an indoor environment. We built our environment using the CoppeliaSim simulator \cite{coppeliaSim}, and the Ivy framework \cite{lenton2021ivy}. Figure \ref{fig:room-adv}  presents the environment which we refer to as \textit{room adversarial} in Table \ref{tab:table-3}. For more detail about each environment, please refer to the supplementary material. We discretize the environments into 3D grid graphs of size $50 \times 50 \times 25$, and randomly remove $5$ sub-graphs of size  $5 \times 5 \times 5$ both during training and testing, this way simulating real-life planning scenarios with random obstacles. The hyperparameter configuration and the specific architecture we utilize are equivalent to Section \ref{sec:experiments-grids}, but with $n = 4$. Likewise, the node features are 3D grid coordinates, and the baselines include supervised learning (SL), $h_{euc}$, A*, and BFS, similarly as in Sections \ref{sec:experiments-grids}, \ref{sec:experiments-diversity}. In Table \ref{tab:table-3} we report the ratio of expanded nodes with respect to $h_{euc}$. 

\xhdr{Video demo} We provide a video demonstration of \alg running in \textit{room adversarial}: \url{https://cutt.ly/eniu5ax}.
\begin{table*}[!htb]
\centering
\resizebox{0.7\columnwidth}{!}{%

\begin{tabular}{cccccccc|c} \toprule
 \textbf{Dataset}                     & \textbf{SL} &\textbf{A*} & $h_{euc}$ & \textbf{BFS} & \textbf{\sail} & \textbf{BFWS} & \textbf{\alg} & \textbf{Shortest path} \\ \midrule
Room simple   & 1.124 & 76.052  & 1.000 & 291.888 & 0.973 & 1.286 & \textbf{0.785} &0.782                 \\ 
Room adversarial  & 2.022 & 67.215   & 1.000 & 238.768 & 0.944 & 1.583 & \textbf{0.895} & 0.853                  \\ \bottomrule
\end{tabular}
}
\caption{Results of \alg in the context of planning for indoor UAV flight. \alg outperforms all baselines in both the \textit{room simple} and \textit{room adversarial} environments while remaining close performance-wise to the optimal number of expansions.}
\label{tab:table-3}
\end{table*}

\xhdr{Discussion} As we can observe in Table \ref{tab:table-3}, \alg outperforms all baselines in both environments. Interestingly, \alg expands only approximately $0.3\%$ more nodes in the simple room than least possible and $4.9\%$ more in the adversarial room case. The same figures for the greedy method ($h_{euc}$) are $27.8\%$  and $17.2 \%$, respectively. These results indicate that \alg is capable of learning planning strategies that are close to optimal in both \textit{simple} and \textit{adversarial} graphs, while the performance of naive heuristics degrades.

\subsection{Runtime Analysis}
We summarize test run-times of different approaches in Appendix G. PHIL runs $57.9\%$ faster than BFWS and $32.2\%$ faster than \sail , and not much slower than traditional A* ($34.7\%$) and $h_{man}$ ($18.3\%$). Although Neural A* is $71.0\%$ faster than PHIL due to the fact that it casts the whole search process into matrix operations on images, it cannot be employed in a generic search setting.

%% file: 005conclusion.tex
\section{Conclusion}

We consider the problem of learning to search for feasible paths in graphs efficiently. We propose a model and a training procedure to learn search heuristics that can be easily deployed across diverse graphs, with tuneable trade-off parameters between constant factors and performance. Our results demonstrate that \alg outperforms current state-of-the-art approaches and can be applied to various graphs with practical use cases.

%% file: 006appendix.tex
\appendix
\newpage

\section{Discussion, Limitations, and Future Work}
\label{sec:future-work}

\paragraph{Injection from nodes and edges to features}  As mentioned in Section \ref{sec:experiments-diversity}, if multiple graph nodes or edges have the same features, our heuristic learning method is challenging to apply. To ensure that \alg has a constant time complexity, we bound the number of neighbouring nodes used for graph convolutions around new fringe nodes, and do not perform any graph convolutions on goal nodes. However, if multiple graph nodes have the same features, or they perhaps do not have any features at all, we may need to perform operations that are not constant in the size of the graph, such as sampling anchor nodes as in position-aware GNN \cite{you2019position}, or collecting more expressive Node2Vec \cite{grover2016node2vec} style features. Since the time complexity of these methods are relatively high (e.g. position-aware GNN's time complexity is $O(|V|^2\log^2|V|)$ while the Dijkstra algorithm only runs in $O(|V|\log|V| + |E|)$), we do not use them unless necessary. For use cases where injections from nodes \& edges to features are hard to guarantee, we encourage practitioners to increase $n$ or potentially perform multiple convolutions on fringe and goal nodes.

\paragraph{Restricted fringe evaluation} As explained in Section \ref{sec:approach}, \alg only evaluates new fringe nodes which are obtained after expanding a node. In practice, this means that once \alg assigns a heuristic value to a node, the value is never updated. While this approach is favorable in terms of the time-complexity of heuristic computations, it does not allow \alg to re-evaluate potentially promising nodes for expansion, based on its updated belief about the POMDP state. We believe that methods that predict the features of promising nodes to expand combined with locality-sensitive hashing or approaches that incorporate node value uncertainty present promising avenues for future work.

\paragraph{Solutions not necessarily optimal} For a best-first search algorithm to find optimal solutions, the used heuristics needs to be \textit{admissible} \cite{russell2002artificial}. In our approach, we do not guarantee the trained heuristics to be \text it{admissible}, which means that when combined with best-first search, \alg would not guarantee optimal final solutions. On the other hand, our approach is concerned with finding \textit{satisficing} solutions as quickly as possible, which is motivated by possible applications in Section \ref{sec:intro}. As in \cite{bhardwaj2017learning}, our learned heuristics can be easily incorporated into a framework such as multi-heuristic A* \cite{aine2016multi}, where any number of inadmissible heuristics can be used with a single admissible heuristic, and the final solution cost sub-optimality is bounded.  An interesting avenue for future work would be to design heuristic learning loss functions that discourage models from over-estimating heuristic values.

\paragraph{Full memory permutation invariance} As noted in Section \ref{sec:approach}, our memory module is invariant to the permutation of nodes in $\VS_{new}$. However, due to how the $\textsc{GRU}$ module is applied, we do not guarantee that that the memory is permutation invariant with respect to the sequence in which nodes are expanded, or equivalently the sequence of $\VS_{new}$ sets. It could be desirable to guarantee such permutation invariance, as the observations are still nodes and edges of a graph, which may not contain sequential inductive biases. Recent work by Cohen \etal \cite{cohen2020regularizing} shows that a simple regularization trick can help efficiently train permutation invariant RNNs. It would be interesting to explore in which cases does full permutation invariance improve \alg's performance.

\paragraph{Directed graphs} As one may notice, most of the examples in this work include graphs that are undirected. The main reason for this is that once we have directed edges in a graph, it may happen that a particular node does not have a path toward the goal, which means that the oracle cost would be effectively undefined. One option for avoiding this issue is adding parallel backward edges during training, ensuring that paths to goals always exist (assuming that the start and goal nodes are parts of the same connected component). This way, \alg is correspondingly penalised for expanding a node that does not immediately lie on a path to the goal.

\paragraph{Ethical considerations} Search algorithms with heuristics can be used within unethical systems. However, our work is not tailored for any particular use cases, and hence we do not believe that it has clear direct negative consequences.

\section{Practical implementation details}

\begin{figure}[htpb]
\centering
\includegraphics[width=0.7\textwidth]{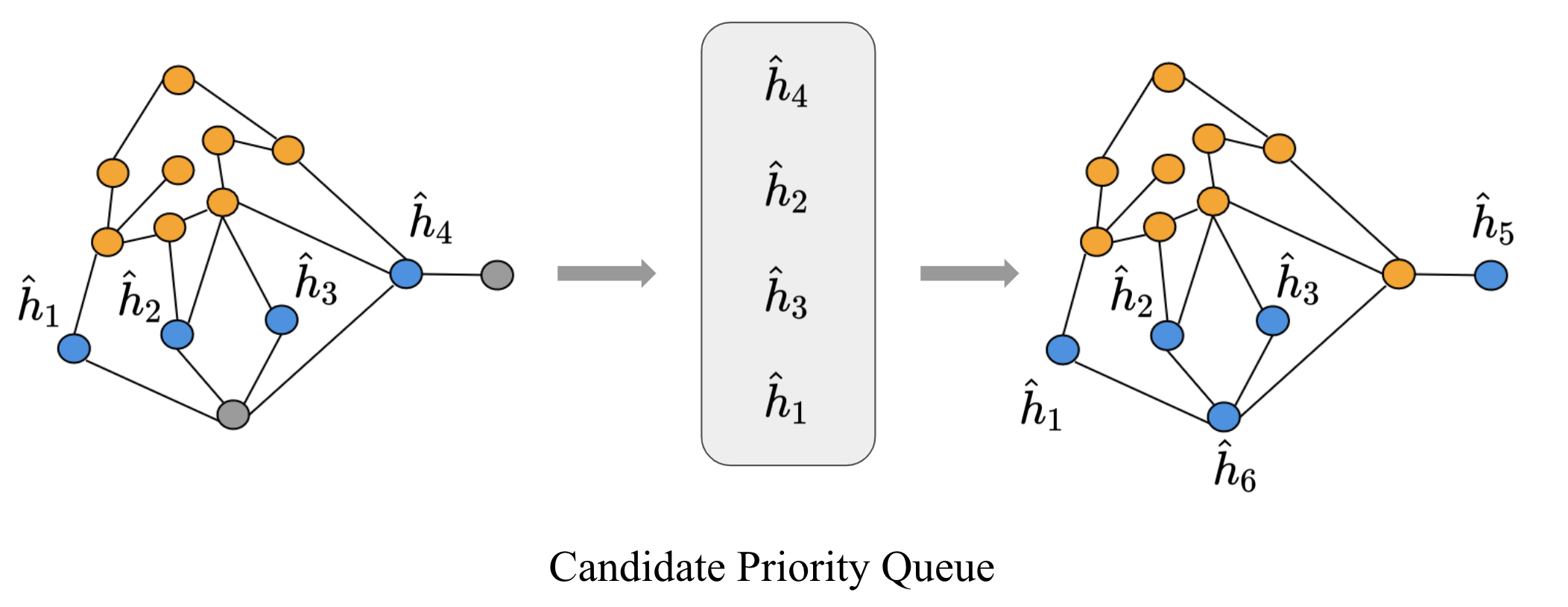}
\caption{PHIL is used as a heuristic to predict the distances of each node, similarly as a heuristic function in best-first search. This figure illustrates a single queue `pop' operation.}
\label{fig:implementation}
\end{figure}

As noted in our abstract, at test time, the heuristic function obtained from \alg can be directly used as a heuristic in an algorithm such as A* or greedy best-first search. In practice, this means that we maintain a priority queue of nodes and distances predicted by the \alg heuristic and greedily expand the nodes which are predicted to be closest to the goal, as seen in Figure \ref{fig:implementation}.

To ensure fast training in Algorithm \ref{alg:il_alg}, we maintain two priority queues, one for \alg and one for the oracle heuristic. On every expansion, we update both the \alg queue and oracle queue. This way, probabilistically blending the two greedy policies comes down to either popping from the \alg queue or the oracle queue.

\section{Training, Baselines, Hyperparameters, and Datasets}
\label{sec:baselines-and-datasets}
\paragraph{Baselines} In our experiments we use a range of both \textit{learning-based} and \textit{classical} \textit{baselines}. Our \textit{baselines} include:

\begin{itemize}
    \item \textbf{Neural A*} \cite{yonetani2020path}: Neural A* learns a differentiable version of A* search on two-dimensional grid graphs. It cannot be straightforwardly adapted to general graphs, which is why we use it as a baseline solely in Section 5.1.
    
    \item \textbf{\sail} \cite{bhardwaj2017learning}: Imitation learning-based approach for learning search heuristics on grid graphs. \sail is adapted in Section 5.2 to not include robotics domain specific features presented in \cite{bhardwaj2017learning}, as these are not possible to compute for general graphs.
    
    \item \textbf{Supervised Learning} (SL): An $\textsc{MLP}$ trained to predict distances between nodes conditioned on node features. In Sections \ref{sec:experiments-diversity}, Sections \ref{sec:experiments-drone} the nodes are sampled from the graph uniformly at random. In Section \ref{sec:experiments-grids}, an oracle policy is rolled out, and data points are collected by sampling random actions during the roll-out as in \sail. This is because in Section \ref{sec:experiments-grids}, the start-goal distribution is not uniform.
    
    \item \textbf{Deep Q Learning}  (QL) \cite{mnih2013playing}: A DQN agent trained to explore the graphs such that $|\VS_{seen}|$ is minimized. The agent receives a negative reward of $-1$ until the goal node is reached in each episode. 
    
    \item \textbf{Cross Entropy Method - Evolutionary Strategies} (CEM) \cite{de2005tutorial}: Derivative free optimization of $h_{\theta}$ using evolutionary strategies. As explained in \cite{bhardwaj2017learning}, the initial population of policies is sampled using a batch size of $40$. Then, each policy is evaluated on $5$ graphs and assigned a score based on the number of expanded nodes for the fitness function. After computing the fitness function, $20\%$ of best-performing policies are selected to be a part of the next population.
    
    \item \textbf{Best-first width search (BFWS)} \cite{lipovetzky2017best}: BFWS adapts greedy best-first search with a generic search history-based novelty metric, which allows it to escape search plateaus and explore relevant nodes. We implement BFWS in experiments from Section 5.1 and 5.3. In Section 5.2, we extend BFWS with boundary-extension features \cite{teichteil2021boundary}, which allows us to apply it to continuous feature spaces. 
    
    \item \textbf{Multi-heuristic A*} (MHA*)\cite{aine2016multi}: Multi-heuristic A* using the Euclidean, Manhattan, and $d_{obs}$ heuristics in a round-robin fashion, where $d_{obs}$ is the distance of the closest uncovered obstacle for a given node.  This baseline is only used in Section \ref{sec:experiments-grids}, where all of these heuristics are well-defined.
    
    \item \textbf{A* search} (A*): A* search algorithm using a Euclidean heuristic function on the node features.
    
    \item \textbf{Greedy best-first search} ($h_{man}, h_{euc}$): Greedy best-first search using Manhattan and Euclidean heuristics, respectively.
    
    \item \textbf{Breadth-first-search} (BFS): Vanilla breadth-first-search without any heuristic.
\end{itemize}

\xhdr{Datasets} Here, we provide more details about the datasets used in Section \ref{sec:experiments-diversity} and Section \ref{sec:experiments-drone}. For more information about datasets used in Section \ref{sec:experiments-grids}, we refer the reader to Bhardwaj and Choudhury \etal \cite{bhardwaj2017learning}.

In our experiments using diverse graphs, as noted in Section \ref{sec:experiments-diversity}, we use the node \& edge features provided in each dataset for training and testing. The three exceptions to this are the PPI, OSMnx datasets, and citation networks. As discussed in Appendix A, we added node labels (i.e., $120$ dimensional label vectors) as features in the PPI dataset because the default node features were not unique. In the OSMnx networks, we did not use all the provided node and edge features, rather only the geographical node coordinates. We further augmented the OSMnx node features with their degree centrality and included Laplacian, modularity, and Bethe Hessian edge features. For citation networks, we only used the first $128$ features to make the search problem more challenging.

In static graphs (such as the OSMnx networks), it may be helpful to augment the graph with expressive features that could be useful for heuristic computations, such as eigenvectors of the graph Laplacian matrix. In non-static graphs, these operations would typically be too expensive to re-compute on each pathfinding attempt. Note that computing more expressive features such as betweenness or percolation centrality have higher time complexities than computing shortest path distances between all pairs of nodes. Due to their time complexity, these features are impractical pre-compute, though they would likely lead to superior search heuristics.

\begin{figure}[!htb]
    \centering
    \begin{minipage}{.4\textwidth}
        \centering
        \includegraphics[width=0.8\linewidth, height=0.15\textheight]{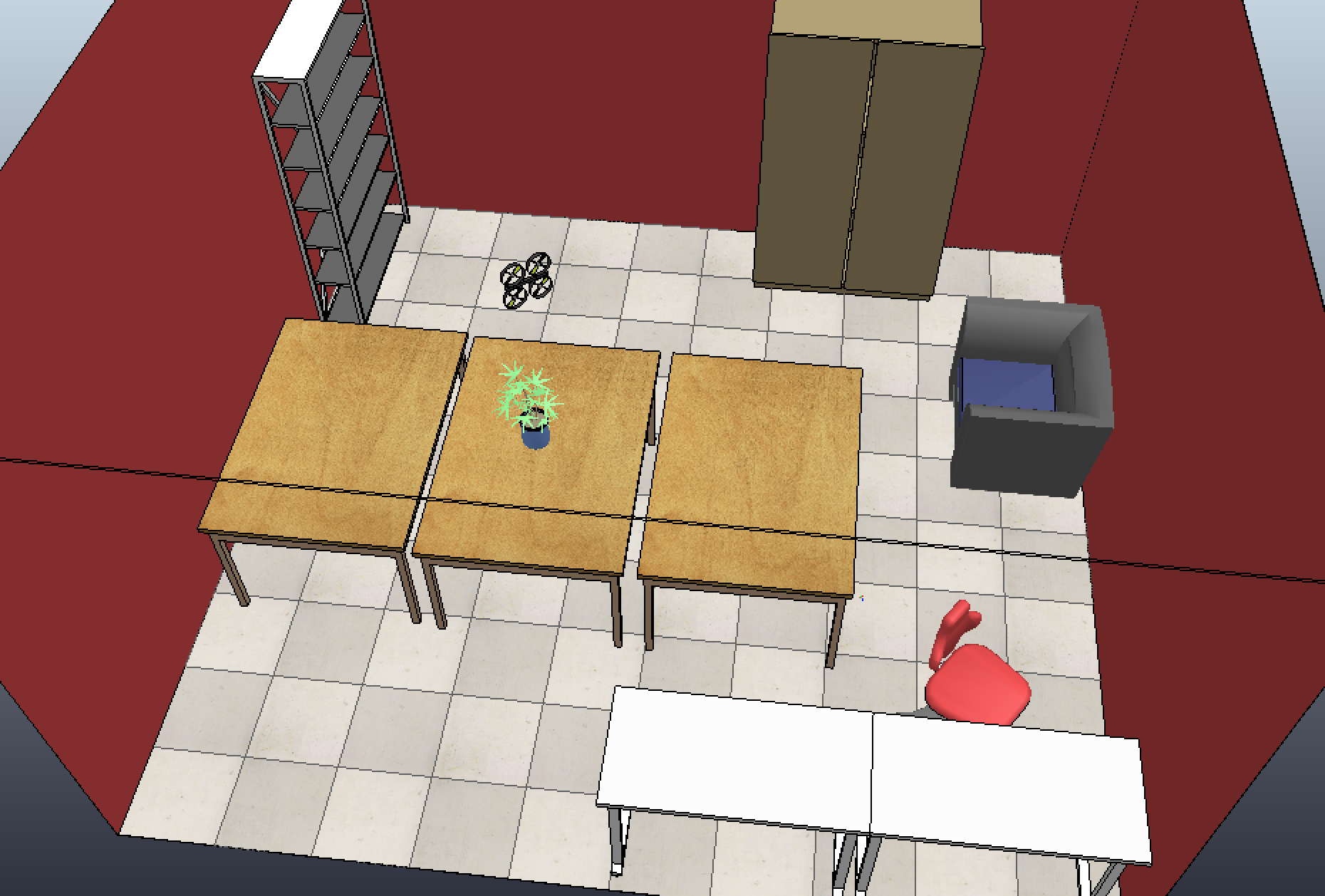}
    \end{minipage}%
    \begin{minipage}{0.4\textwidth}
        \centering
        \includegraphics[width=0.8\linewidth, height=0.15\textheight]{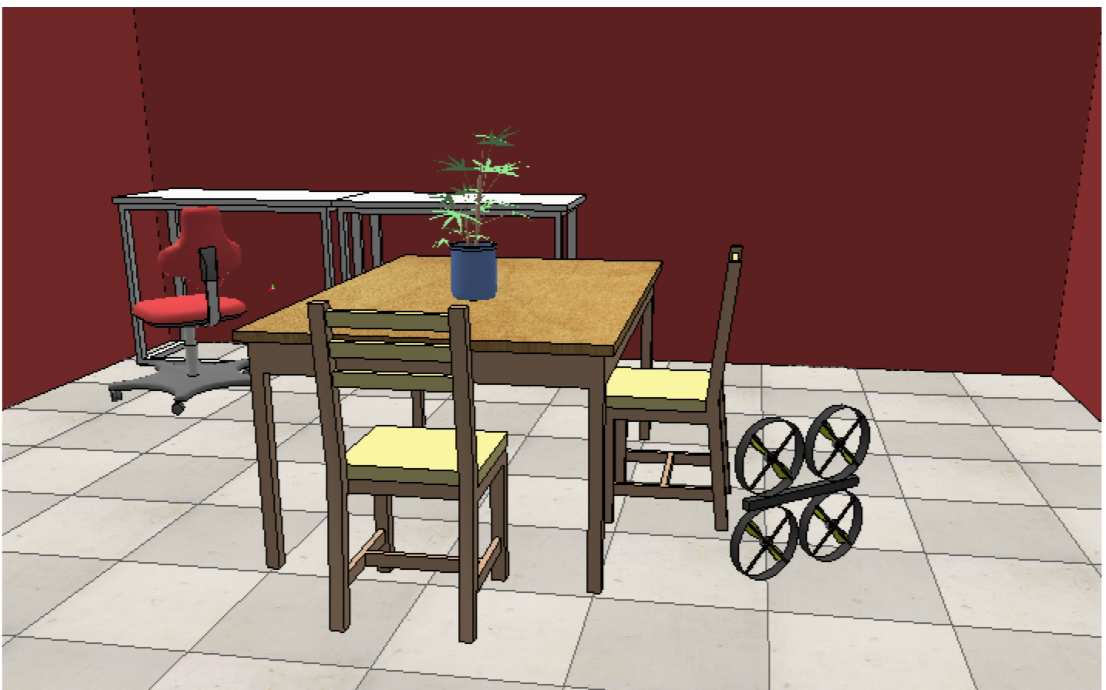}
    \end{minipage}

\caption{We present two CoppeliaSim environments used for our UAV flight experiments. On the left, we have the \textit{adversarial room environment}, on the right, we present the \textit{room simple} environment.}
\label{fig:rooms}
\end{figure}

In Section \ref{sec:experiments-drone},  we built our environment using the CoppeliaSim simulator \cite{coppeliaSim}, and the Ivy framework by Lenton \etal \cite{lenton2021ivy}. Figure \ref{fig:rooms} presents the environments which we refer to as \textit{room adversarial} and \textit{room simple} in Table \ref{tab:table-3}. The \textit{room simple} environment is equivalent to the Ivy drone demo environment --- a single room with a table surrounded by chairs in the middle of the room and various furniture close to the walls. The main difference between \textit{room simple} and \textit{room adversarial} is that \textit{room simple} only contains a single table in the middle of the room. In contrast, in \textit{room} \textit{adversarial}, three tables span to a wall, this was creating a bottleneck for naive heuristics. Note that the drones are not allowed to fly under furniture to make the environment more challenging.

\paragraph{Hyperparameters in diverse graphs} With regards to the diverse graph experiments (Section \ref{sec:experiments-diversity}), there are in-practice three parameters that we can tune for each dataset: $n$, $t_\tau$ $T$. Recall that the hyperparameter $n$ is the maximal size of the 1-hop neighborhood around the new fringe nodes sampled during training. We found that a rule of thumb of setting $n$ to the minimum of $8$ and the average node degree in the given graph dataset worked reasonably well. In practice, $n$ should be tuned to optimize the search effort \& wall-clock performance trade-off. In terms of $T$, it is advisable to set it such that the algorithm can reach target nodes during training. Hence, we found that setting $T$ roughly to the largest graph diameter is a suitable choice. In terms of $t_\tau$, we train each sequence using the stored old rolled-in states, and for larger graphs ($>10,000$ nodes), we use $t_\tau = 128$, while for smaller graphs (between $10,000$ and $5,000$ nodes), we use $t_\tau = 64$, and otherwise we use a rule of thumb of taking $t_\tau \approx T * 0.2$.

\paragraph{Grid graphs \& drone flight hyperparameters} In our grid graph experiments (Section \ref{sec:experiments-grids}), we use $T = 256$,  $t_{\tau} = 32$, $\beta_{0} = 0.7$, $n = 8$. Finally, in our drone flight experiments, we use the same configuration as in the grid graph experiments. The only differences are that we set $n = 4$,  $t_{\tau} = 64$, and we randomize start and goal nodes both during training and testing.

\paragraph{Optimization} \alg is generally trained using the Adam optimizer with a learning rate of $0.001$ and a batch size of $32$. Once we sample $t \sim \mathcal{U}(0, \dots, T-t_\tau)$ for a roll-in, it can happen that the target is reached in less than $t + t_\tau$ steps. In such cases, we continue the sequence until we reach $t_\tau$ steps or all the graph nodes are explored, in which case we end the episode. Another approach would be to end episodes once the goal node is reached. In practice, we did not find significant performance differences between the two methods.

\section{Ablation studies}
\label{sec:ablation}
\begin{figure}[!htb]
\minipage{0.32\textwidth}
  \centering
  \frame{\includegraphics[width=0.4\linewidth]{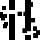}}
\endminipage\hfill
\minipage{0.32\textwidth}
  \centering
  \frame{\includegraphics[width=0.4\linewidth]{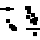}}
\endminipage\hfill
\minipage{0.32\textwidth}%
  \centering
  \frame{\includegraphics[width=0.4\linewidth]{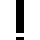}}
\endminipage

\caption{This figure presents the down-scaled datasets used for the ablation experiments. We have \textit{Gaps+Forest} (left), \textit{Forest} (centre), and \textit{Alternating gaps} (right).}
\label{fig:envs}
\end{figure}

For the ablation studies, we use three, $4$-connected versions of down-scaled datasets provided in Bhardwaj \etal \cite{bhardwaj2017learning}: \textit{Gaps+Forest}, \textit{Forest}, and \textit{Alternating gaps}. We downscale each dataset $5\times$, to dimensions $40 \times 40$. While the \textit{Gaps+Forest} dataset presents a more challenging environment with multiple planning bottlenecks and obstacles, \textit{Forest} and \textit{Alternating gaps} are simpler environments with more straightforward heuristics. Figure \ref{fig:envs} present example graphs from the down-scaled datasets \textit{Gaps+Forest}, \textit{Forest}, and \textit{Alternating gaps}.

In each ablation experiment, unless stated otherwise, we use a batch size of $8$, the Adam optimizer with a learning rate of $0.01$, and up to $20$ roll-in steps. The $\textsc{MLP}$ in \alg has $3$ layers of width $128$, \textit{LeakyReLU} activations, and the memory state has $d=64$ dimensions. 2D grid coordinates are used as node features, similarly as in Section \ref{sec:experiments-grids}. We report the number of explored nodes with respect to A* or with respect to the optimal number of expansions (i.e., the shortest path distances between nodes). All experiments are performed across $3$ random seeds, and standard deviations are used for error bars.

\subsection{Zeroed-out states vs. Rolled-in states}
\label{sec:zeronero}

There is a trade-off between using rolled-in states for downstream training using backpropagation through time (i.e., storing $z_t$ after each roll-in) and using zeroed-out states (i.e., storing zeroed-out initial states). Namely, past rolled-in states $z_t$ are out-of-distribution for optimized versions of $h_\theta$, but \alg may use these embeddings for inferring initial node distances because $z_t$ contains information about the rolled-in graph. On the other hand, zeroed-out states are always in-distribution for $h_\theta$, but the algorithm is constrained to start from an initial state of $\vec{\mathbf{0}}$ in regions where this may never be the case at test time. Our goal is to gain insight into when one method is preferable over to the other in this ablation.

\begin{figure}[htpb]
\centering
\includegraphics[width=0.6\textwidth]{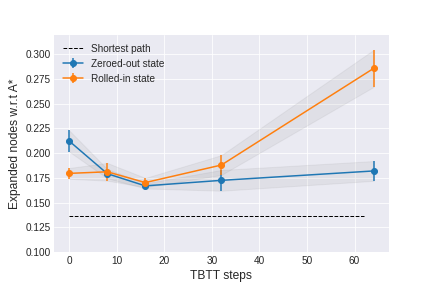}
\caption{This plot illustrates the performance of \alg with zeroed-out initial states vs. stored rolled-in initial states during training across multiple TBTT steps.}
\label{fig:rolled-vs-zero}
\end{figure}

We train two versions of \alg, one with zeroed-out initial states for TBTT on each trajectory (\alg-zero), and one with initial states stored from roll-ins of the past versions of \alg (\alg-roll). Both versions are trained for $t_\tau = 0, 8, 16, 32, 64$ TBTT steps using the \textit{Gaps+Forest} graph dataset. Figure \ref{fig:rolled-vs-zero} illustrates the performance progress across TBTT steps of both of these approaches. Firstly, we observe that up to around $t_\tau = 16$ steps; both approaches positively benefit from performing more backpropagation steps through time, reinforcing that the memory module brings overall performance benefits. Further, we may see the performance difference of \alg-roll and \alg-zero at $0$ steps. In this region, \alg-roll outperforms \alg-zero by about $3\%$, which is following the intuition of the rolled-in states containing useful information about the embeddings of rolled-in graphs. Secondly, we can notice that the performance of \alg-zero plateaus after $16$ steps. This plateau suggests that the graph may not contain useful information for inferring node distances beyond $16$ backpropagated steps. Finally, the performance of \alg-roll decreases much steeper than that of \alg-zero once it starts deteriorating after $16$ steps, which could mean that there is some form of error compounding once \alg makes predictions from a wrongly initialised state during training.

\paragraph{Conclusion} The main takeaway of this ablation experiment is that if one would like to perform only a few TBTT steps (i.e., low $t_\tau$), using rolled-in states will likely provide some performance benefits. On the other hand, as  $t_\tau$ increases, it is preferable to use zeroed-out initial states. Practitioners can determine this choice on a per-problem basis.

\subsection{Effect of GNN choice} 

In this ablation we assess what effects do different graph neural networks have on the performance of \alg. Namely, we train the \alg-zero from the previous ablation, using $t_\tau = 16$, $n=4$ and five different GNNs: GAT (Veličković \etal \cite{velivckovic2017graph}), MPNN (\textit{max}), MPNN (\textit{sum}) (Gilmer \etal \cite{gilmer2017neural}), DeeperGCN (\textit{softmax}), and DeeperGCN (\textit{power}) (Li \etal \cite{li2020deepergcn}). We report the ratio of explored nodes with respect to A* in Table \ref{tab:table-4}.

\begin{table*}[!htb]
\centering
\resizebox{0.7\columnwidth}{!}{%

\begin{tabular}{cccc} \toprule
 \textbf{GNN}   & \textbf{Alternating gaps} & \textbf{Forest} & \textbf{Gaps+Forest} \\ \midrule
GAT   & 0.077 ± 0.0110 & 0.065 ± 0.0013  & \textbf{0.154 ± 0.0112} \\ 
MPNN (\textit{max})   & 0.071 ± 0.0834 & \textbf{0.064 ± 0.0004}  & 0.158 ± 0.0143  \\ 
MPNN (\textit{sum})   & 0.095 ± 0.0487 & 0.07 ± 0.0085  & 0.187 ± 0.0135  \\ 
DeeperGCN (\textit{softmax})   & \textbf{0.069 ± 0.0008} & \textbf{0.064 ± 0.0030}  & 0.164 ± 0.0113 \\ 
DeeperGCN (\textit{power})  & 0.076 ± 0.0027 & 0.07 ± 0.0101  & 0.19 ± 0.0037   \\ \bottomrule
\end{tabular}
}
\caption{This table presents the results obtained using different graph convolutions in the three graph datasets from the ablation study. We can observe that maximisation-based aggregation performs better on average, while attention can provide performance benefits in the challenging \textit{Gaps+Forest} graphs.}
\label{tab:table-4}
\end{table*}

\begin{wrapfigure}[13]{r}{0.4\textwidth}
  \centering
  \includegraphics[width=0.5\linewidth]{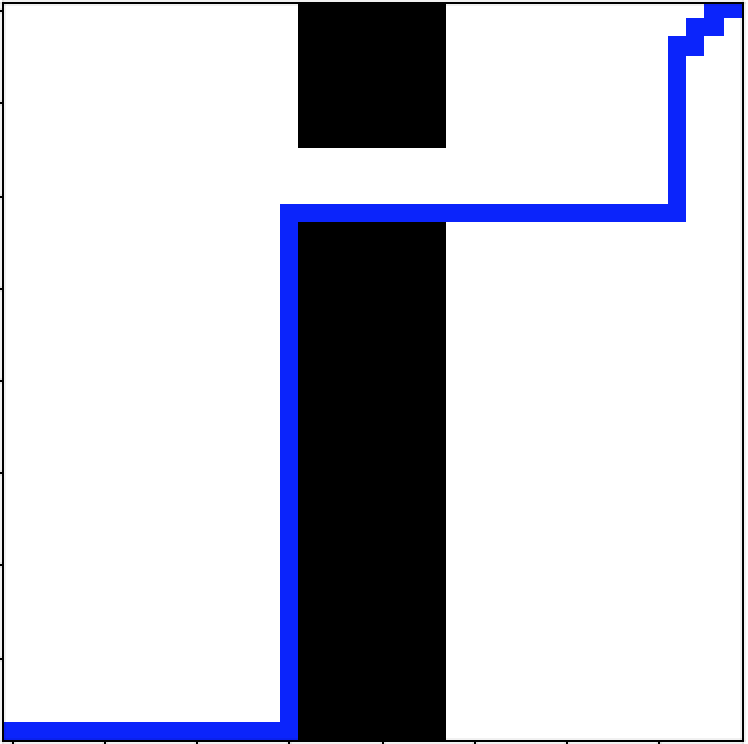}
    \caption{This figure illustrates \alg using the DeeperGCN (\textit{softmax}) GNN learning the optimal heuristic strategy in the \textit{Alternating gaps} dataset.}
  \label{fig:optimal-strategy}
\end{wrapfigure}

In general, we find that maximisation-based aggregation approaches tend to perform better than other approaches by $1.57 \%$ on average. Note that this comparison only includes the MPNN and DeeperGCN GNNs. DeeperGCN (\textit{softmax}) achieves the best results on both the \textit{Alternating gaps} and \textit{Forest} datasets. Further, using DeeperGCN (\textit{softmax}), \alg learned the optimal strategy for finding the goal in the \textit{Alternating gaps} dataset, which is to follow the path along the bottleneck wall, as seen in Figure \ref{fig:optimal-strategy}.

\begin{figure}[htp]

\centering
\includegraphics[width=.2\textwidth]{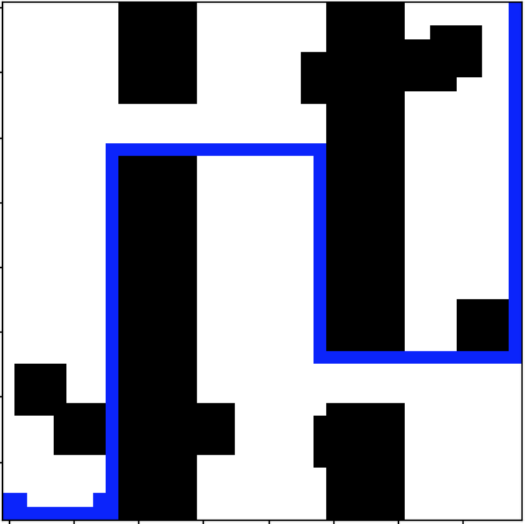}\hfill
\includegraphics[width=.2\textwidth]{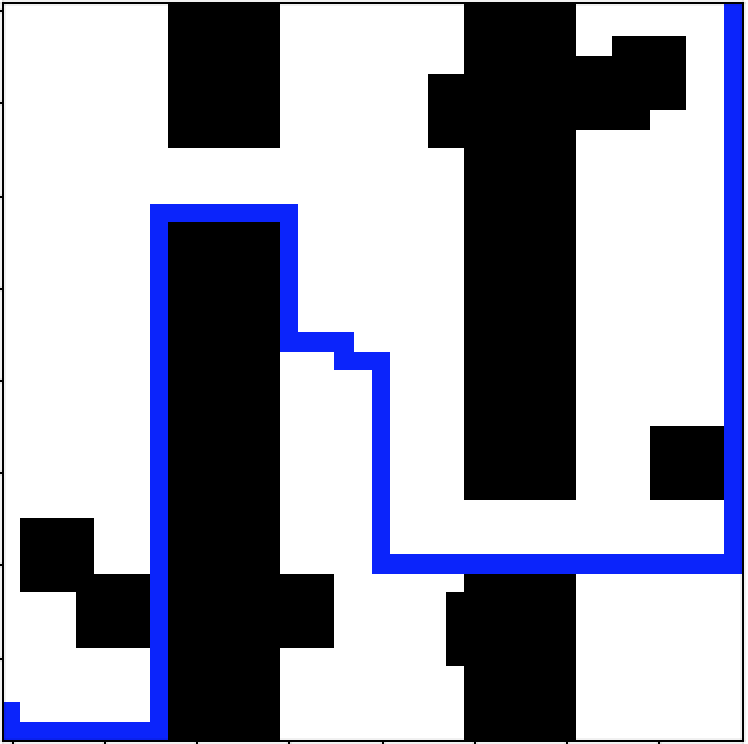}\hfill
\includegraphics[width=.2\textwidth]{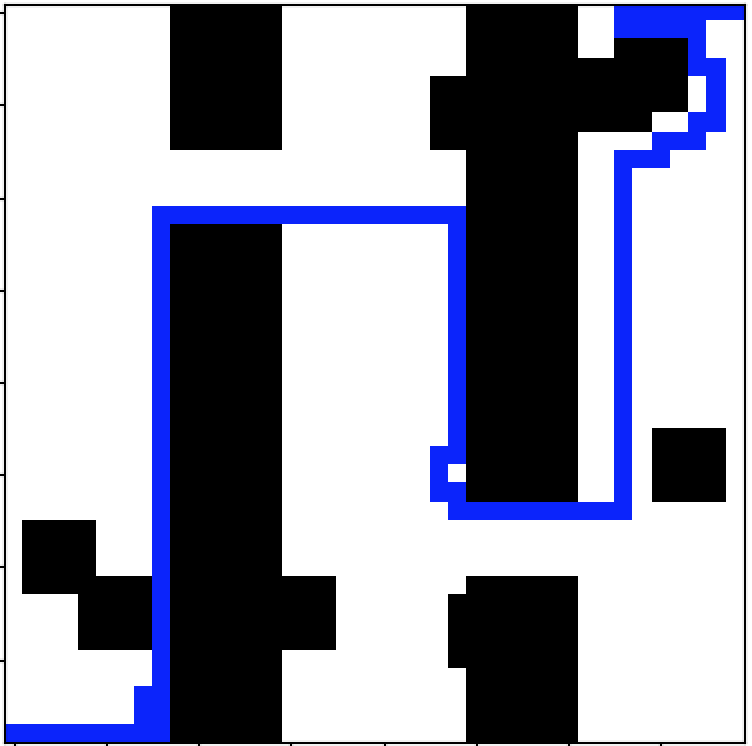}

\caption{We compare the solutions of \alg trained with a GAT (left), DeeperGCN (\textit{softmax}) (middle), and MPNN (\textit{sum}) (right). Results demonstrate that MPNN (\textit{sum}) makes several redundant expansions while the other two methods expand paths close to optimally.}
\label{fig:figure-3convs}
\end{figure}

GAT outperforms other approaches on the \textit{Gaps+Forest} dataset, which is also the most complex of the three datasets. This finding suggests that forms of attention could be useful for learning heuristics in more complex graphs. In Figure \ref{fig:figure-3convs}, we can see a side-by-side comparison of GAT, DeeperGCN (\textit{softmax}), and MPNN (\textit{sum}) in the \textit{Gaps+Forest} graph dataset. We may observe that the GAT-based and DeeperGCN-based heuristics find strategies that are close to optimal, while MPNN (\textit{sum}) performs a similar strategy, but with slightly more expansions.

\paragraph{Conclusion} The GNN ablation demonstrates that maximisation-based aggregation performs better, with an average reduction of explored nodes by $1.57 \%$. This finding is consistent with Veličković \etal \cite{velivckovic2019neural}, where GNNs are applied to execute graph algorithms. While the problem solved by  Veličković \etal \cite{velivckovic2019neural} is different, its nature is similar: train a GNN to \textit{select} which node to explore next in order to imitate a reference graph algorithm. In \alg, scores are assigned to nodes rather than nodes being selected, but the downstream operation is node selection. Further, these experiments also suggest that attention can be helpful in more complex graphs, with GAT outperforming other approaches in the \textit{Gaps+Forest} graphs.

\subsection{Increasing neighborhood size} 
\begin{figure}[htpb]
\centering
\includegraphics[width=0.6\textwidth]{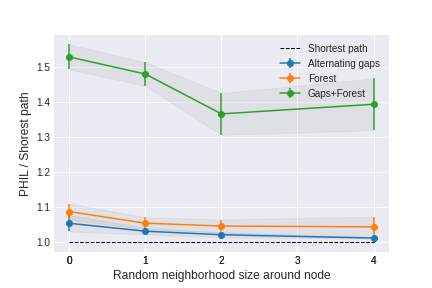}
\caption{This plot illustrates the performance of \alg with increasing $n = |\mathcal{N}_i|$ through $0, 1, 2, 4$ for each evaluated fringe node $i$.}
\label{fig:sampling_extent}
\end{figure}

As explained in the approach (Section \ref{sec:approach}), for each new fringe node we uniformly sample an $n \in \mathbb{N}_{\geq 0}$ bounded neighborhood of nodes, which we then use for graph convolutions. We hypothesized that with increasing $n$, the performance of \alg will improve. In Figure \ref{fig:sampling_extent}, we validate this hypothesis on \textit{Alternating gaps}, \textit{Forest}, and \textit{Gaps+Forest} datasets, by gradually setting $n = 0, 1, 2, 4$. We train \alg using a DeeperGCN (\textit{softmax}) graph convolution, $t_\tau = 16$, and with otherwise the same set of hyperparameters as in the \textit{Zeroed-out states vs. Rolled-in states} (Appendix \ref{sec:zeronero}) experiment. In Figure \ref{fig:sampling_extent}, we report the explored node ratio of each method with respect to the optimal number of explored nodes, that is, the shortest paths between start \& goal pairs.

\paragraph{Conclusion} In Figure \ref{fig:sampling_extent}, we may observe that on all datasets the performance of \alg improves with increasing $n = |\mathcal{N}_i|$ until $n = 2$, after which it plateaus. These results suggest that \alg can benefit from additional nodes sampled from the neighbourhoods of evaluated fringe nodes, even if this sampling is performed uniformly at random.

\subsection{Increasing memory capacity}

\begin{figure}[!htpb]
\centering
\includegraphics[width=0.6\textwidth]{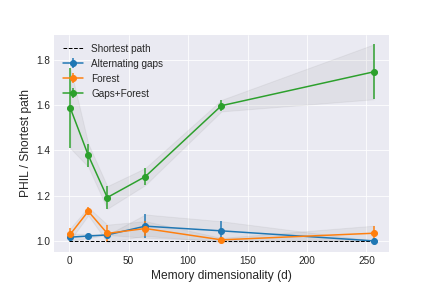}
\caption{This plot illustrates the performance of \alg with increasing $d$ through $1, 16, 32, 64, 128, 256$ on the three ablation datasets.}
\label{fig:memory_capacity}
\end{figure}

In the final ablation experiment, we consider what effects changing the capacity of memory ($d$) has on the overall performance of \alg. We alter $d = 1, 16, 32, 64, 128, 256$ across all datasets. In Figure \ref{fig:memory_capacity}, the ratio of explored nodes is presented with respect to the shortest path length, which is the optimal baseline.

Focusing on the \textit{Gaps+Forest} dataset, in Figure \ref{fig:memory_capacity} we can observe that the performance of \alg generally improves until about $d = 32$ by approximately $40 \%$ with respect to $d=1$, after which it starts getting worse. Hence, we can conclude that additional memory capacity can be helpful for \alg to learn representations better suited for inferring distances of newly added fringe nodes. Note that we trained \alg for a fixed number of iterations ($N = 36$) in all cases, which means that the decrease in performance after $d = 32$ could also be due to the $\textsc{GRU}$ module having more parameters to optimise, which may take longer to converge. However, it could also easily be that the memory module starts overfitting to samples in the aggregated dataset during training. In the case of the simpler \textit{Alternating gaps} and \textit{Forest} datasets, the differences between different amounts of memory capacity are marginal. These findings are supported by approaches such as \sail \cite{bhardwaj2017learning} achieving good performances in simpler environments. By implication, performance decrease in the \textit{Gaps+Forest} dataset for larger values of $d$ is more likely due to overfitting than optimisation difficulties.

\paragraph{Conclusion} Additional memory capacity is crucial for \alg to learn useful representations in more challenging graphs, while the importance of additional memory decreases as the graphs are simpler. However, a certain amount of overfitting is observed for larger values of $d$, which means that $d$ should be tuned on a per problem basis.

\newpage
\section{Architecture}

\begin{figure}[!htpb]
    \centering
    \includegraphics[width=0.75\textwidth]{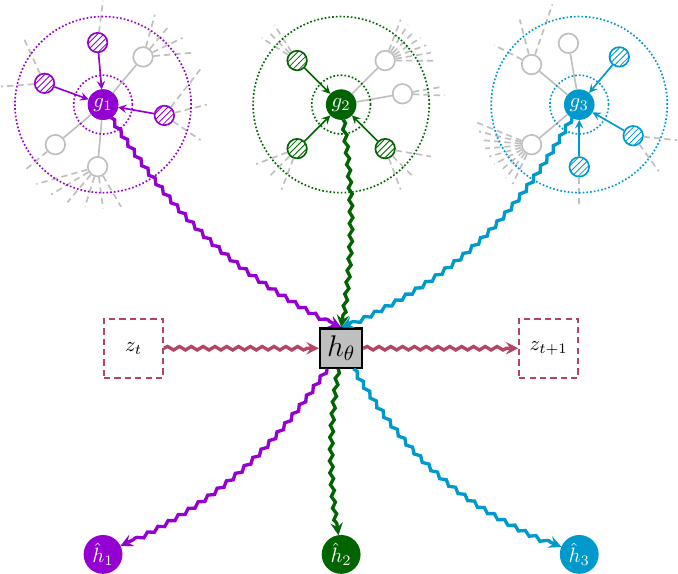}
    \caption{This figure illustrates the computation performed in a single forward pass of $h_\theta$. In the case of this figure, we would have $|\VS_{new}| = n = 3$, with $g_i$ representing the convolved embeddings, and $\hat{h}_i$ the output heuristic values. Horizontally, we illustrate the update of memory $z_t$ to $z_{t+1}$. This figure is adapted from \cite{figurespetar}.
    }
    \label{fig:inference}
\end{figure}

\section{Qualitative Examples}
\begin{figure}[!htpb]
    \centering
    \includegraphics[width=0.9\columnwidth]{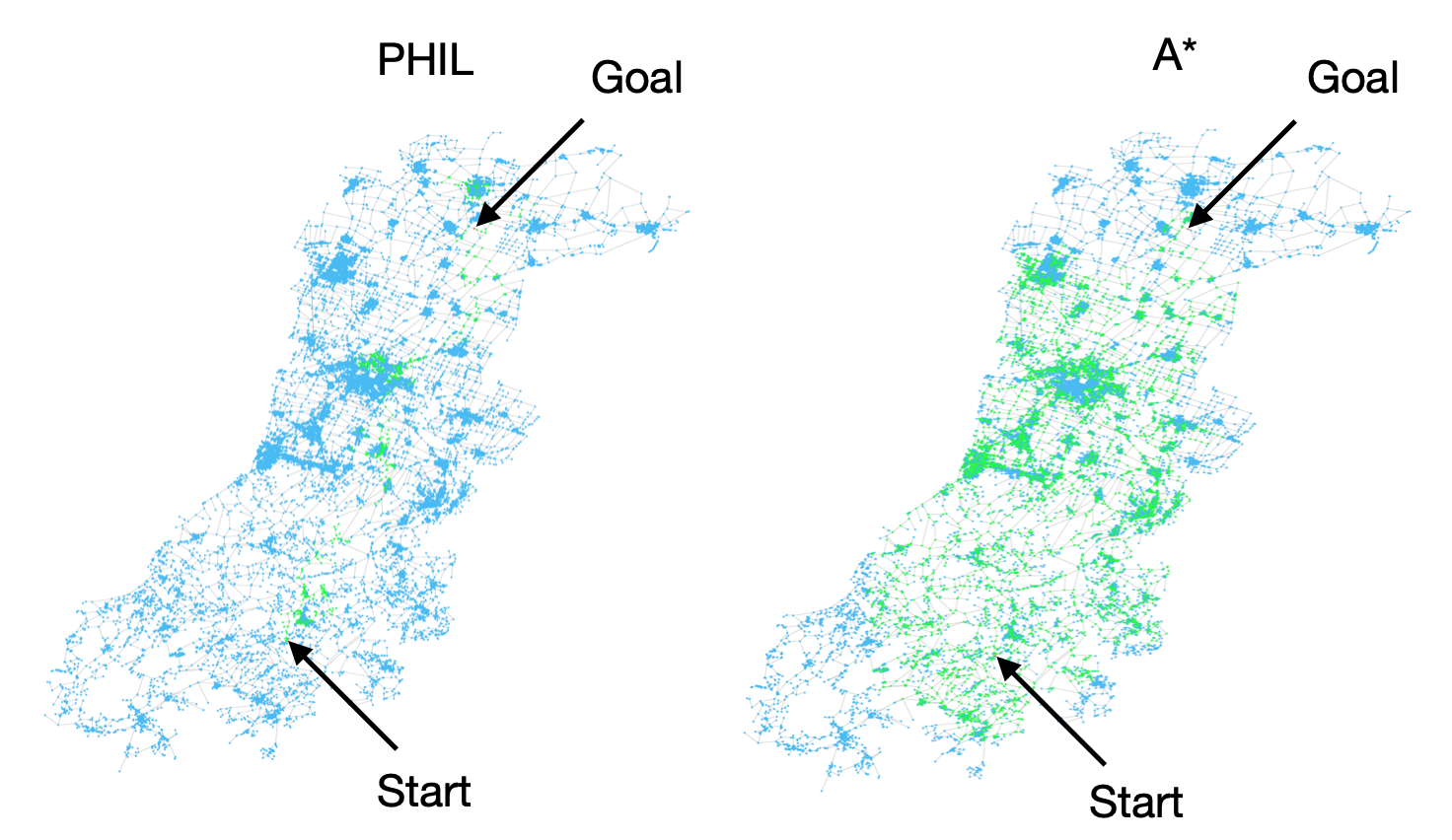}
    \caption{This figure illustrates the road network of Modena, Italy. We contrast the search effort of \alg (left) and A* (right). We can observe that \alg expands (shown in green) considerably fewer nodes searching for the goal $v_g$.}
    \label{fig:modena}
\end{figure}

\begin{figure}[!htpb]
    \centering
    \includegraphics[width=0.9\columnwidth]{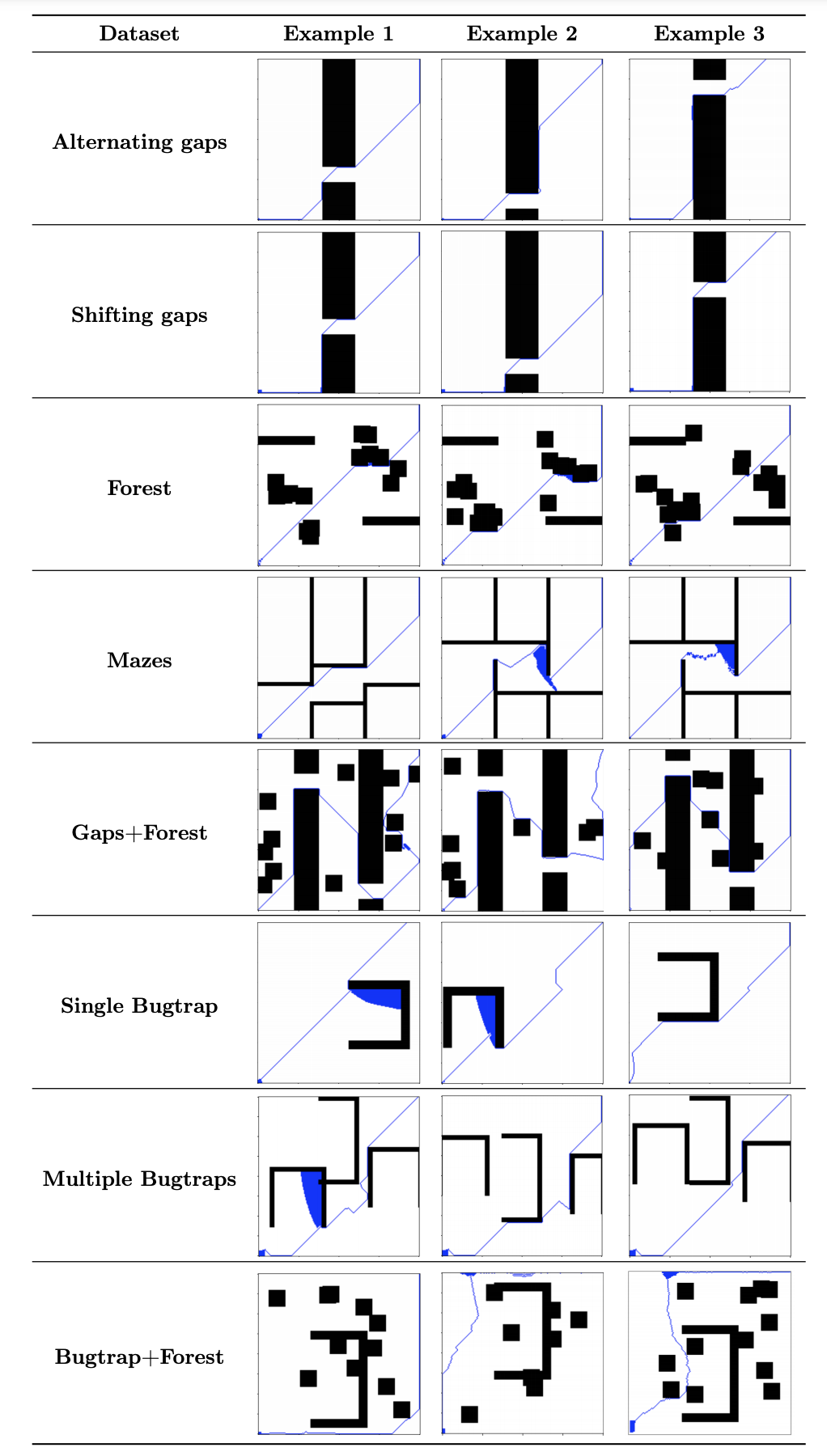}
    \caption{This figure presents representative examples of \alg runs in datasets from Section \ref{sec:experiments-grids}}
    \label{fig:grid}
\end{figure}

\section{Runtime Analysis}
\label{sec:runtime}
\begin{figure}[!htpb]
    \centering
    \includegraphics[width=0.9\columnwidth]{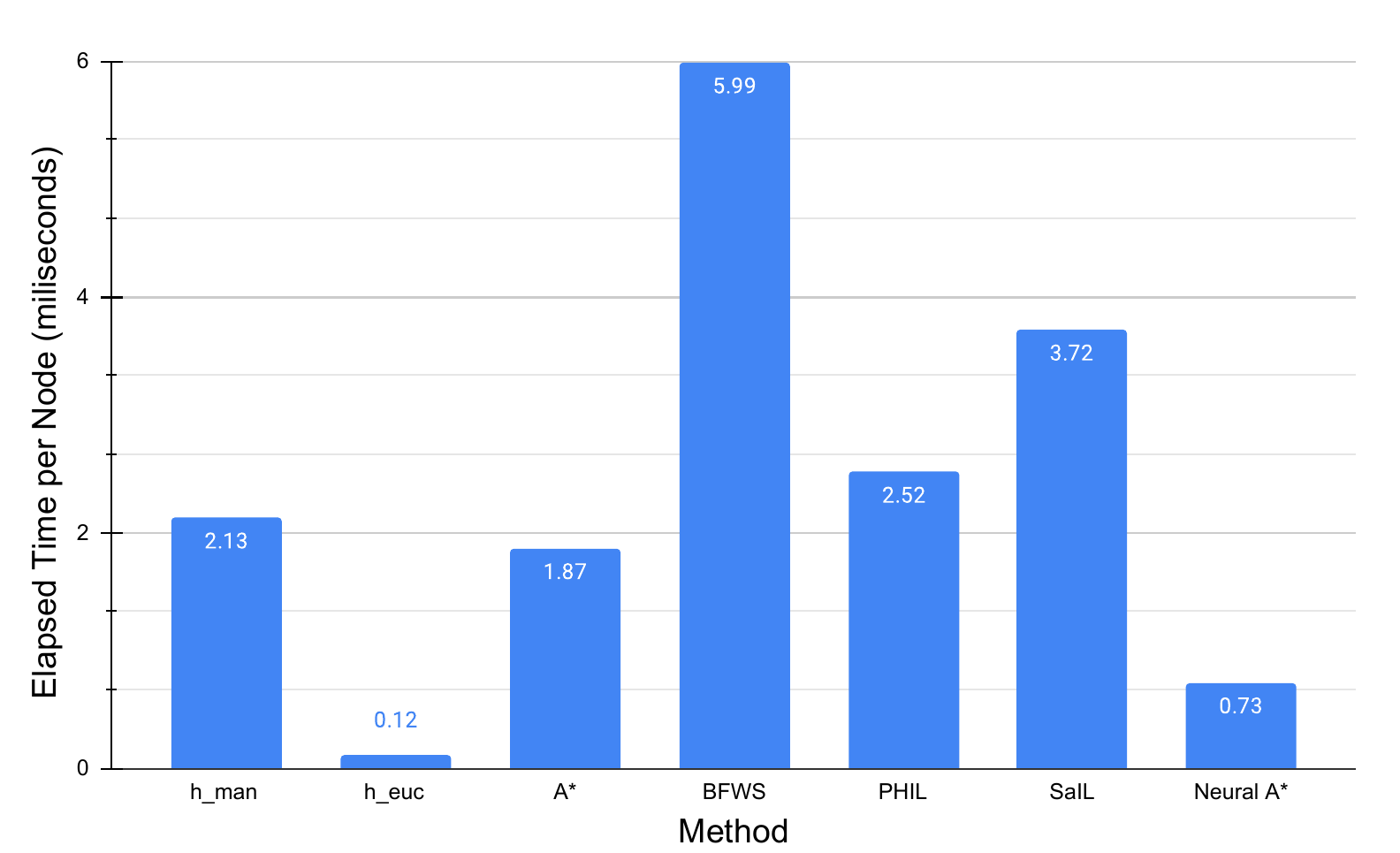}
    \caption{Timing experiments on grid search datasets \cite{bhardwaj2017learning}.}
    \label{fig:runtime}
\end{figure}

We evaluate runtimes of 7 approaches during inference on all the datasets in Section \ref{sec:experiments-grids}. Each approach's runtime are averaged by number of nodes it visited. The runtime experiments are conducted on a machine with a RTX5000 GPU and a Intel Xeon 5222 CPU.

%% file: log_2022.bbl
\begin{thebibliography}{67}
\providecommand{\natexlab}[1]{#1}
\providecommand{\url}[1]{\texttt{#1}}
\expandafter\ifx\csname urlstyle\endcsname\relax
  \providecommand{\doi}[1]{doi: #1}\else
  \providecommand{\doi}{doi: \begingroup \urlstyle{rm}\Url}\fi

\bibitem[Bhardwaj et~al.(2017)Bhardwaj, Choudhury, and
  Scherer]{bhardwaj2017learning}
Mohak Bhardwaj, Sanjiban Choudhury, and Sebastian Scherer.
\newblock Learning heuristic search via imitation.
\newblock In \emph{Conference on Robot Learning}, 2017.

\bibitem[Chen et~al.(2020)Chen, Li, Dai, and Song]{chen2020retro}
Binghong Chen, Chengtao Li, Hanjun Dai, and Le~Song.
\newblock Retro*: Learning retrosynthetic planning with neural guided a*
  search.
\newblock In \emph{ICML}, 2020.

\bibitem[Gebser et~al.(2013)Gebser, Kaufmann, Romero, Otero, Schaub, and
  Wanko]{gebser2013domain}
Martin Gebser, Benjamin Kaufmann, Javier Romero, Ram{\'{o}}n Otero, Torsten
  Schaub, and Philipp Wanko.
\newblock Domain-specific heuristics in answer set programming.
\newblock In \emph{AAAI}, 2013.

\bibitem[Mac et~al.(2016)Mac, Copot, Tran, and De~Keyser]{mac2016heuristic}
Thi~Thoa Mac, Cosmin Copot, Duc~Trung Tran, and Robin De~Keyser.
\newblock Heuristic approaches in robot path planning: A survey.
\newblock In \emph{Robotics and Autonomous Systems}, 2016.

\bibitem[Sharma and Goolsbey(2017)]{sharma2017identifying}
Abhishek Sharma and Keith~M. Goolsbey.
\newblock Identifying useful inference paths in large commonsense knowledge
  bases by retrograde analysis.
\newblock In \emph{{AAAI}}, 2017.

\bibitem[Yeh et~al.(2012)Yeh, Yeh, Arias, and Soo]{yeh2012pathway}
Cheng-Yu Yeh, Hsiang-Yuan Yeh, Carlos~Roberto Arias, and Von-Wun Soo.
\newblock Pathway detection from protein interaction networks and gene
  expression data using color-coding methods and a* search algorithms.
\newblock In \emph{The Scientific World booktitle}, 2012.

\bibitem[Pearl(1984)]{pearl1984heuristics}
Judea Pearl.
\newblock \emph{Heuristics: intelligent search strategies for computer problem
  solving}.
\newblock 1984.

\bibitem[Khalidi et~al.(2020)Khalidi, Gujarathi, and Saha]{khalidi2020t}
Danish Khalidi, Dhaval Gujarathi, and Indranil Saha.
\newblock T*: A heuristic search based path planning algorithm for temporal
  logic specifications.
\newblock In \emph{{ICRA}}, 2020.

\bibitem[Adabala and Ajanovic(2020)]{adabala2020multi}
Bhargav Adabala and Zlatan Ajanovic.
\newblock A multi-heuristic search-based motion planning for autonomous
  parking.
\newblock In \emph{30th International Conference on Automated Planning and
  Scheduling: Planning and Robotics Workshop}, 2020.

\bibitem[Xie et~al.(2012)Xie, Nakhost, and M{\"u}ller]{xie2012planning}
Fan Xie, Hootan Nakhost, and Martin M{\"u}ller.
\newblock Planning via random walk-driven local search.
\newblock In \emph{Twenty-Second International Conference on Automated Planning
  and Scheduling}, 2012.

\bibitem[Xie et~al.(2014)Xie, M{\"u}ller, and Holte]{xie2014adding}
Fan Xie, Martin M{\"u}ller, and Robert Holte.
\newblock Adding local exploration to greedy best-first search in satisficing
  planning.
\newblock In \emph{AAAI}, 2014.

\bibitem[Lipovetzky and Geffner(2017)]{lipovetzky2017best}
Nir Lipovetzky and Hector Geffner.
\newblock Best-first width search: Exploration and exploitation in classical
  planning.
\newblock In \emph{AAAI}, 2017.

\bibitem[Teichteil-K{\"o}nigsbuch et~al.(2021)Teichteil-K{\"o}nigsbuch,
  Ramirez, and Lipovetzky]{teichteil2021boundary}
Florent Teichteil-K{\"o}nigsbuch, Miquel Ramirez, and Nir Lipovetzky.
\newblock Boundary extension features for width-based planning with simulators
  on continuous-state domains.
\newblock In \emph{Proceedings of the Twenty-Ninth International Conference on
  International Joint Conferences on Artificial Intelligence}, 2021.

\bibitem[Bonet and Geffner()]{bonet2001planning}
Blai Bonet and H{\'e}ctor Geffner.
\newblock Planning as heuristic search.
\newblock pages 5--33.

\bibitem[Zhu and Givan(2003)]{zhu2003landmark}
Lin Zhu and Robert Givan.
\newblock Landmark extraction via planning graph propagation.
\newblock 2003.

\bibitem[Richter and Westphal(2010)]{richter2010lama}
Silvia Richter and Matthias Westphal.
\newblock The lama planner: Guiding cost-based anytime planning with landmarks.
\newblock 2010.

\bibitem[Sutskever(2013)]{sutskever2013training}
Ilya Sutskever.
\newblock \emph{Training recurrent neural networks}.
\newblock University of Toronto, Toronto, Canada, 2013.

\bibitem[Ross and Bagnell(2014)]{ross2014reinforcement}
Stephane Ross and J~Andrew Bagnell.
\newblock Reinforcement and imitation learning via interactive no-regret
  learning.
\newblock In \emph{arXiv preprint arXiv:1406.5979}, 2014.

\bibitem[Choudhury et~al.(2017)Choudhury, Kapoor, Ranade, Scherer, and
  Dey]{choudhury2017adaptive}
Sanjiban Choudhury, Ashish Kapoor, Gireeja Ranade, Sebastian Scherer, and
  Debadeepta Dey.
\newblock Adaptive information gathering via imitation learning.
\newblock 2017.

\bibitem[Arfaee et~al.(2011)Arfaee, Zilles, and Holte]{arfaee2011learning}
Shahab~Jabbari Arfaee, Sandra Zilles, and Robert~C Holte.
\newblock Learning heuristic functions for large state spaces.
\newblock In \emph{Artificial Intelligence}, 2011.

\bibitem[{\'u}s~Virseda et~al.(2013){\'u}s~Virseda, Borrajo, and
  Alc{\'a}zar]{us2013learning}
Jes {\'u}s~Virseda, Daniel Borrajo, and Vidal Alc{\'a}zar.
\newblock Learning heuristic functions for cost-based planning.
\newblock In \emph{Planning and Learning}, 2013.

\bibitem[Wilt and Ruml(2015)]{wilt2015building}
Christopher~Makoto Wilt and Wheeler Ruml.
\newblock Building a heuristic for greedy search.
\newblock In \emph{SOCS}, 2015.

\bibitem[Garrett et~al.(2016)Garrett, Kaelbling, and
  Lozano{-}P{\'{e}}rez]{garrett2016learning}
Caelan~Reed Garrett, Leslie~Pack Kaelbling, and Tom{\'{a}}s
  Lozano{-}P{\'{e}}rez.
\newblock Learning to rank for synthesizing planning heuristics.
\newblock In \emph{IJCAI}, 2016.

\bibitem[Thayer et~al.(2011)Thayer, Dionne, and Ruml]{thayer2011learning}
Jordan Thayer, Austin Dionne, and Wheeler Ruml.
\newblock Learning inadmissible heuristics during search.
\newblock In \emph{Proceedings of the International Conference on Automated
  Planning and Scheduling}, 2011.

\bibitem[Kim and An(2020)]{KimA20}
Soonkyum Kim and Byungchul An.
\newblock Learning heuristic a*: efficient graph search using neural network.
\newblock In \emph{{ICRA}}, 2020.

\bibitem[Ariki and Narihira(2019)]{ariki2019fully}
Yuka Ariki and Takuya Narihira.
\newblock Fully convolutional search heuristic learning for rapid path
  planners.
\newblock In \emph{arXiv preprint arXiv:1908.03343}, 2019.

\bibitem[Terasawa et~al.(2020)Terasawa, Ariki, Narihira, Tsuboi, and
  Nagasaka]{terasawa20203d}
Ryo Terasawa, Yuka Ariki, Takuya Narihira, Toshimitsu Tsuboi, and Kenichiro
  Nagasaka.
\newblock 3d-cnn based heuristic guided task-space planner for faster motion
  planning.
\newblock In \emph{{ICRA}}, 2020.

\bibitem[Yonetani et~al.(2021)Yonetani, Taniai, Barekatain, Nishimura, and
  Kanezaki]{yonetani2020path}
Ryo Yonetani, Tatsunori Taniai, Mohammadamin Barekatain, Mai Nishimura, and
  Asako Kanezaki.
\newblock Path planning using neural a* search.
\newblock In \emph{ICML}, 2021.

\bibitem[Archetti et~al.(2021)Archetti, Cannici, and
  Matteucci]{archetti2021neural}
Alberto Archetti, Marco Cannici, and Matteo Matteucci.
\newblock Neural weighted a*: Learning graph costs and heuristics with
  differentiable anytime a.
\newblock 2021.

\bibitem[Khalil et~al.(2017)Khalil, Dai, Zhang, Dilkina, and
  Song]{NIPS2017_d9896106}
Elias~B. Khalil, Hanjun Dai, Yuyu Zhang, Bistra Dilkina, and Le~Song.
\newblock Learning combinatorial optimization algorithms over graphs.
\newblock In \emph{{NeurIPS}}, 2017.

\bibitem[Li et~al.(2018)Li, Chen, and Koltun]{NEURIPS2018_8d3bba74}
Zhuwen Li, Qifeng Chen, and Vladlen Koltun.
\newblock Combinatorial optimization with graph convolutional networks and
  guided tree search.
\newblock In \emph{{NeurIPS}}, 2018.

\bibitem[Karalias and Loukas(2020)]{NEURIPS2020_49f85a9e}
Nikolaos Karalias and Andreas Loukas.
\newblock Erdos goes neural: an unsupervised learning framework for
  combinatorial optimization on graphs.
\newblock In \emph{{NeurIPS}}, 2020.

\bibitem[Silver and Veness(2010)]{silver2010monte}
David Silver and Joel Veness.
\newblock Monte-carlo planning in large pomdps.
\newblock In \emph{{NeurIPS}}, 2010.

\bibitem[Guez et~al.(2018)Guez, Weber, Antonoglou, Simonyan, Vinyals, Wierstra,
  Munos, and Silver]{guez2018learning}
Arthur Guez, Theophane Weber, Ioannis Antonoglou, Karen Simonyan, Oriol
  Vinyals, Daan Wierstra, R{\'{e}}mi Munos, and David Silver.
\newblock Learning to search with mctsnets.
\newblock In \emph{{ICML}}, 2018.

\bibitem[Deac et~al.(2020)Deac, Veli{\v{c}}kovi{\'c}, Milinkovi{\'c}, Bacon,
  Tang, and Nikoli{\'c}]{deac2020xlvin}
Andreea Deac, Petar Veli{\v{c}}kovi{\'c}, Ognjen Milinkovi{\'c}, Pierre-Luc
  Bacon, Jian Tang, and Mladen Nikoli{\'c}.
\newblock Xlvin: executed latent value iteration nets.
\newblock In \emph{arXiv preprint arXiv:2010.13146}, 2020.

\bibitem[Karkus et~al.(2017)Karkus, Hsu, and Lee]{karkus2017qmdp}
P{\'{e}}ter Karkus, David Hsu, and Wee~Sun Lee.
\newblock Qmdp-net: Deep learning for planning under partial observability.
\newblock In \emph{{NeurIPS}}, 2017.

\bibitem[Tamar et~al.(2016)Tamar, Levine, Abbeel, Wu, and
  Thomas]{tamar2016value}
Aviv Tamar, Sergey Levine, Pieter Abbeel, Yi~Wu, and Garrett Thomas.
\newblock Value iteration networks.
\newblock In \emph{{NeurIPS}}, 2016.

\bibitem[Pascanu et~al.(2017)Pascanu, Li, Vinyals, Heess, Buesing,
  Racani{\`e}re, Reichert, Weber, Wierstra, and Battaglia]{pascanu2017learning}
Razvan Pascanu, Yujia Li, Oriol Vinyals, Nicolas Heess, Lars Buesing, Sebastien
  Racani{\`e}re, David Reichert, Th{\'e}ophane Weber, Daan Wierstra, and Peter
  Battaglia.
\newblock Learning model-based planning from scratch.
\newblock In \emph{arXiv preprint arXiv:1707.06170}, 2017.

\bibitem[Racani{\`{e}}re et~al.(2017)Racani{\`{e}}re, Weber, Reichert, Buesing,
  Guez, Rezende, Badia, Vinyals, Heess, Li, Pascanu, Battaglia, Hassabis,
  Silver, and Wierstra]{weber2017imagination}
S{\'{e}}bastien Racani{\`{e}}re, Theophane Weber, David~P. Reichert, Lars
  Buesing, Arthur Guez, Danilo~Jimenez Rezende, Adri{\`{a}}~Puigdom{\`{e}}nech
  Badia, Oriol Vinyals, Nicolas Heess, Yujia Li, Razvan Pascanu, Peter~W.
  Battaglia, Demis Hassabis, David Silver, and Daan Wierstra.
\newblock Imagination-augmented agents for deep reinforcement learning.
\newblock In \emph{{NeurIPS}}, 2017.

\bibitem[Daum{\'e} et~al.(2009)Daum{\'e}, Langford, and Marcu]{daume2009search}
Hal Daum{\'e}, John Langford, and Daniel Marcu.
\newblock Search-based structured prediction.
\newblock In \emph{Machine learning}, 2009.

\bibitem[Chang et~al.(2015)Chang, Krishnamurthy, Agarwal, III, and
  Langford]{chang2015learning}
Kai{-}Wei Chang, Akshay Krishnamurthy, Alekh Agarwal, Hal~Daum{\'{e}} III, and
  John Langford.
\newblock Learning to search better than your teacher.
\newblock In \emph{ICML}, 2015.

\bibitem[Sun et~al.(2017)Sun, Venkatraman, Gordon, Boots, and
  Bagnell]{sun2017deeply}
Wen Sun, Arun Venkatraman, Geoffrey~J. Gordon, Byron Boots, and J.~Andrew
  Bagnell.
\newblock Deeply aggrevated: differentiable imitation learning for sequential
  prediction.
\newblock In \emph{{ICML}}, 2017.

\bibitem[Laskey et~al.(2017)Laskey, Lee, Fox, Dragan, and
  Goldberg]{laskey2017dart}
Michael Laskey, Jonathan Lee, Roy Fox, Anca Dragan, and Ken Goldberg.
\newblock Dart: Noise injection for robust imitation learning.
\newblock In \emph{Conference on robot learning}, 2017.

\bibitem[Sun et~al.(2018)Sun, Bagnell, and Boots]{sun2018truncated}
Wen Sun, J.~Andrew Bagnell, and Byron Boots.
\newblock Truncated horizon policy search: combining reinforcement learning
  {\&} imitation learning.
\newblock In \emph{{ICLR}}, 2018.

\bibitem[Cho et~al.(2014)Cho, van Merri{\"e}nboer, Gulcehre, Bahdanau,
  Bougares, Schwenk, and Bengio]{cho-etal-2014-learning}
Kyunghyun Cho, Bart van Merri{\"e}nboer, Caglar Gulcehre, Dzmitry Bahdanau,
  Fethi Bougares, Holger Schwenk, and Yoshua Bengio.
\newblock Learning phrase representations using {rnn} encoder{--}decoder for
  statistical machine translation.
\newblock In \emph{{EMNLP}}, 2014.

\bibitem[Gilmer et~al.(2017)Gilmer, Schoenholz, Riley, Vinyals, and
  Dahl]{gilmer2017neural}
Justin Gilmer, Samuel~S. Schoenholz, Patrick~F. Riley, Oriol Vinyals, and
  George~E. Dahl.
\newblock Neural message passing for quantum chemistry.
\newblock In \emph{ICML}, 2017.

\bibitem[Li et~al.(2020)Li, Xiong, Thabet, and Ghanem]{li2020deepergcn}
Guohao Li, Chenxin Xiong, Ali Thabet, and Bernard Ghanem.
\newblock Deepergcn: All you need to train deeper gcns.
\newblock In \emph{arXiv preprint arXiv:2006.07739}, 2020.

\bibitem[You et~al.(2019)You, Ying, and Leskovec]{you2019position}
Jiaxuan You, Rex Ying, and Jure Leskovec.
\newblock Position-aware graph neural networks.
\newblock In \emph{{ICML}}, 2019.

\bibitem[Hu et~al.(2020)Hu, Fey, Zitnik, Dong, Ren, Liu, Catasta, and
  Leskovec]{hu2020open}
Weihua Hu, Matthias Fey, Marinka Zitnik, Yuxiao Dong, Hongyu Ren, Bowen Liu,
  Michele Catasta, and Jure Leskovec.
\newblock Open graph benchmark: Datasets for machine learning on graphs.
\newblock In \emph{{NeurIPS}}, 2020.

\bibitem[Sen et~al.(2008)Sen, Namata, Bilgic, Getoor, Galligher, and
  Eliassi-Rad]{sen2008collective}
Prithviraj Sen, Galileo Namata, Mustafa Bilgic, Lise Getoor, Brian Galligher,
  and Tina Eliassi-Rad.
\newblock Collective classification in network data.
\newblock In \emph{AI magazine}, 2008.

\bibitem[Shchur et~al.(2018)Shchur, Mumme, Bojchevski, and
  G{\"u}nnemann]{shchur2018pitfalls}
Oleksandr Shchur, Maximilian Mumme, Aleksandar Bojchevski, and Stephan
  G{\"u}nnemann.
\newblock Pitfalls of graph neural network evaluation.
\newblock In \emph{arXiv preprint arXiv:1811.05868}, 2018.

\bibitem[Zitnik and Leskovec(2017)]{zitnik2017predicting}
Marinka Zitnik and Jure Leskovec.
\newblock Predicting multicellular function through multi-layer tissue
  networks.
\newblock In \emph{Bioinformatics}, 2017.

\bibitem[Morris et~al.(2020)Morris, Kriege, Bause, Kersting, Mutzel, and
  Neumann]{morris2020tudataset}
Christopher Morris, Nils~M Kriege, Franka Bause, Kristian Kersting, Petra
  Mutzel, and Marion Neumann.
\newblock Tudataset: A collection of benchmark datasets for learning with
  graphs.
\newblock In \emph{arXiv preprint arXiv:2007.08663}, 2020.

\bibitem[Boeing(2017)]{boeing2017osmnx}
Geoff Boeing.
\newblock Osmnx: New methods for acquiring, constructing, analyzing, and
  visualizing complex street networks.
\newblock In \emph{Computers, Environment and Urban Systems}, 2017.

\bibitem[Allamanis(2019)]{allamanis2019adverse}
Miltiadis Allamanis.
\newblock The adverse effects of code duplication in machine learning models of
  code.
\newblock In \emph{ACM SIGPLAN International Symposium on New Ideas, New
  Paradigms, and Reflections on Programming and Software}, 2019.

\bibitem[Wu et~al.(2018)Wu, Ramsundar, Feinberg, Gomes, Geniesse, Pappu,
  Leswing, and Pande]{wu2018moleculenet}
Zhenqin Wu, Bharath Ramsundar, Evan~N Feinberg, Joseph Gomes, Caleb Geniesse,
  Aneesh~S Pappu, Karl Leswing, and Vijay Pande.
\newblock Moleculenet: a benchmark for molecular machine learning.
\newblock In \emph{Chemical science}, 2018.

\bibitem[Rohmer et~al.(2013)Rohmer, Singh, and Freese]{coppeliaSim}
E.~Rohmer, S.~P.~N. Singh, and M.~Freese.
\newblock Coppeliasim (formerly v-rep): a versatile and scalable robot
  simulation framework.
\newblock In \emph{IROS}, 2013.

\bibitem[Lenton et~al.(2021)Lenton, Pardo, Falck, James, and
  Clark]{lenton2021ivy}
Daniel Lenton, Fabio Pardo, Fabian Falck, Stephen James, and Ronald Clark.
\newblock Ivy: Templated deep learning for inter-framework portability.
\newblock In \emph{arXiv preprint arXiv:2102.02886}, 2021.

\bibitem[Grover and Leskovec(2016)]{grover2016node2vec}
Aditya Grover and Jure Leskovec.
\newblock node2vec: Scalable feature learning for networks.
\newblock In \emph{{ACM} {SIGKDD}}, 2016.

\bibitem[Russell and Norvig(2002)]{russell2002artificial}
Stuart Russell and Peter Norvig.
\newblock \emph{Artificial intelligence: a modern approach}.
\newblock 2002.

\bibitem[Aine et~al.(2016)Aine, Swaminathan, Narayanan, Hwang, and
  Likhachev]{aine2016multi}
Sandip Aine, Siddharth Swaminathan, Venkatraman Narayanan, Victor Hwang, and
  Maxim Likhachev.
\newblock Multi-heuristic a*.
\newblock In \emph{The International booktitle of Robotics Research}, 2016.

\bibitem[Cohen{-}Karlik et~al.(2020)Cohen{-}Karlik, David, and
  Globerson]{cohen2020regularizing}
Edo Cohen{-}Karlik, Avichai~Ben David, and Amir Globerson.
\newblock Regularizing towards permutation invariance in recurrent models.
\newblock In \emph{NeurIPS}, 2020.

\bibitem[Mnih et~al.(2013)Mnih, Kavukcuoglu, Silver, Graves, Antonoglou,
  Wierstra, and Riedmiller]{mnih2013playing}
Volodymyr Mnih, Koray Kavukcuoglu, David Silver, Alex Graves, Ioannis
  Antonoglou, Daan Wierstra, and Martin Riedmiller.
\newblock Playing atari with deep reinforcement learning.
\newblock In \emph{{NeurIPS} Deep Learning Workshop}, 2013.

\bibitem[De~Boer et~al.(2005)De~Boer, Kroese, Mannor, and
  Rubinstein]{de2005tutorial}
Pieter-Tjerk De~Boer, Dirk~P Kroese, Shie Mannor, and Reuven~Y Rubinstein.
\newblock A tutorial on the cross-entropy method.
\newblock In \emph{Annals of operations research}, 2005.

\bibitem[Velickovic et~al.(2018)Velickovic, Cucurull, Casanova, Romero,
  Li{\`{o}}, and Bengio]{velivckovic2017graph}
Petar Velickovic, Guillem Cucurull, Arantxa Casanova, Adriana Romero, Pietro
  Li{\`{o}}, and Yoshua Bengio.
\newblock Graph attention networks.
\newblock In \emph{{ICLR}}, 2018.

\bibitem[Velickovic et~al.(2020)Velickovic, Ying, Padovano, Hadsell, and
  Blundell]{velivckovic2019neural}
Petar Velickovic, Rex Ying, Matilde Padovano, Raia Hadsell, and Charles
  Blundell.
\newblock Neural execution of graph algorithms.
\newblock In \emph{{ICLR}}, 2020.

\bibitem[Velickovic()]{figurespetar}
Petar Velickovic.
\newblock Tikz.
\newblock \url{https://github.com/PetarV-/TikZ}, last accessed on 01/6/21.

\end{thebibliography}
